\documentclass[acmlarge]{acmart}

\AtBeginDocument{%
  \providecommand\BibTeX{{%
    \normalfont B\kern-0.5em{\scshape i\kern-0.25em b}\kern-0.8em\TeX}}}

\usepackage{soul}
\usepackage[normalem]{ulem}
\setstcolor{red}
\sethlcolor{lime}
\usepackage{graphbox} 
\usepackage[export]{adjustbox} 
\usepackage{caption}
\usepackage{booktabs} 
\usepackage{amsmath}

\def\etal{\emph{et al}.}

\newcommand{\figref}[1]{Figure~\ref{#1}}

\newcommand\redsout{\bgroup\markoverwith{\textcolor{red}{\rule[0.5ex]{2pt}{0.4pt}}}\ULon}

\usepackage{booktabs} 
\usepackage{booktabs}
\usepackage{graphicx}
\usepackage{xcolor}
\usepackage{url}
\usepackage{siunitx}
\usepackage{tabularx}
\usepackage{colortbl}
\usepackage{textcomp, color, soul}
\usepackage{color,soul}
\usepackage{amsmath, wrapfig}
\usepackage{tikz}
\usepackage{ctable}
\usepackage{tikzsymbols}
\usepackage{textcomp}

\usepackage{booktabs} 
\usepackage{courier}
\usepackage{booktabs}
\usepackage{graphicx}
\usepackage{xcolor}
\usepackage{multirow}

\usepackage{tcolorbox}

\usepackage[colorinlistoftodos]{todonotes}

\usepackage{fontawesome}    

\usepackage{url}
\usepackage{tabularx}
\usepackage{textcomp, color, soul, xcolor}
\usepackage{color,soul}
\usepackage{amsmath, wrapfig}
\usepackage{tikz}
\usepackage{ctable}
\usepackage{capt-of}
\usepackage{subcaption, balance}
\usepackage{balance,bm}
\usepackage [english]{babel}
\usepackage{url}
\usepackage [autostyle, english = american]{csquotes}

\definecolor{linkColor}{RGB}{6,125,233}
\definecolor{green}{rgb}{0.0, 0.65, 0.31}
\definecolor{bleudefrance}{rgb}{0.19, 0.55, 0.91}
\definecolor{ceruleanblue}{rgb}{0.16, 0.32, 0.75}
\definecolor{mediumblue}{rgb}{0.0, 0.0, 0.8}
\definecolor{grey}{HTML}{969696}
\definecolor{violet}{HTML}{756bb1}
\definecolor{dgrey}{HTML}{01665e}
\definecolor{lgrey}{HTML}{5ab4ac}
\definecolor{dgreen}{HTML}{005a32}
\definecolor{purple}{HTML}{ae017e}

\definecolor{editCol}{HTML}{0000FF}
\definecolor{maskCol}{HTML}{c51b7d}
\definecolor{lrColor}{HTML}{8856a7}
\definecolor{trColor}{HTML}{d01c8b}
\definecolor{ctColor}{HTML}{4dac26}
\definecolor{brickred}{HTML}{f03b20}
\definecolor{improveCol}{HTML}{253494}
\definecolor{worsenCol}{HTML}{d7191c}
\definecolor{violet}{HTML}{6a51a3}

\definecolor{rubinered}{HTML}{CE0058}
\definecolor{midnightblue}{HTML}{191970}

\newcommand*{\textlabel}[2]{%
  \edef\@currentlabel{#1}
  \phantomsection
  #1\label{#2}
}

\colorlet{tableheadcolor}{gray!25} 
\colorlet{tablerowcolor}{gray!15} 
\colorlet{tablerowcolor2}{gray!12} 
\definecolor{ao(english)}{rgb}{0.0, 0.5, 0.0}

\newcommand{\edit}[1]{{\textcolor{editCol}{#1}}}

\begin{document}

\title{IMUTube: Automatic Extraction of Virtual on-body Accelerometry from Video for Human Activity Recognition}
\author{Hyeokhyen Kwon}
\authornote{Both authors contributed equally to this research.}
\email{hyeokhyen@gatech.edu}
\affiliation{%
  \institution{School of Interactive Computing, Georgia Institute of Technology}
  \city{Atlanta, GA}
  \country{USA}
}

\author{Catherine Tong}
\authornotemark[1]
\email{eu.tong@cs.ox.ac.uk}
\affiliation{%
  \institution{Department of Computer Science, University of Oxford}
  \country{UK}
  }

\author{Harish Haresamudram}
\email{harishkashyap@gatech.edu}
\affiliation{%
  \institution{School of Electrical and Computer Engineering, Georgia Institute of Technology}
  \city{Atlanta, GA}
  \country{USA}
 }
 
\author{Yan Gao}
\email{yan.gao@keble.ox.ac.uk}
\affiliation{%
  \institution{Department of Computer Science, University of Oxford}
  \country{UK}
}

\author{Gregory D. Abowd}
\email{abowd@gatech.edu}
\affiliation{%
  \institution{School of Interactive Computing, Georgia Institute of Technology}
  \city{Atlanta, GA}
  \country{USA}
}

\author{Nicholas D. Lane}
\email{nicholas.lane@cs.ox.ac.uk}
\affiliation{%
  \institution{Department of Computer Science, University of Oxford}
  \country{UK}
}

\author{Thomas Pl{\"o}tz}
\email{thomas.ploetz@gatech.edu}
\affiliation{
  \institution{School of Interactive Computing, Georgia Institute of Technology}
  \city{Atlanta, GA}
  \country{USA}
}
\renewcommand{\shortauthors}{Kwon and Tong, et al.}

\begin{abstract}


The lack of large-scale, labeled data sets impedes progress in developing robust and generalized predictive models for on-body sensor-based human activity recognition (HAR). Labeled data in human activity recognition is scarce and hard to come by, as sensor data collection is expensive, and the annotation is time-consuming and error-prone. To address this problem, we introduce IMUTube, an automated processing pipeline that integrates existing computer vision and signal processing techniques to convert videos of human activity into virtual streams of IMU data. These virtual IMU streams represent accelerometry at a wide variety of locations on the human body. We show how the virtually-generated IMU data improves the performance of a variety of models on known HAR datasets.  Our initial results are very promising, but the greater promise of this work lies in a collective approach by the computer vision, signal processing, and activity recognition communities to extend this work in ways that we outline. This should lead to on-body, sensor-based HAR becoming yet another success story in large-dataset breakthroughs in recognition. 
\end{abstract}


\setcopyright{acmlicensed}
\acmJournal{IMWUT}
\acmYear{2020} \acmVolume{4} \acmNumber{3} \acmArticle{87} \acmMonth{9} \acmPrice{15.00}\acmDOI{10.1145/3411841}


\maketitle

\pagebreak
\section{INTRODUCTION}
On-body sensor-based human activity recognition (HAR) is widely utilized for behavioral analysis, such as user authentication, healthcare, and tracking everyday activities~\cite{scholl2015wearables,Zhang12ubicomp,Chavarriaga13prl,Bachlin10titb,liaqat2019wearbreathing}.
Regardless of its utility, the HAR field has yet to experience significant improvements in recognition accuracy, in contrast to the breakthroughs in other fields, such as speech recognition~\cite{hinton2012deep}, natural language processing~\cite{devlin2018bert}, and computer vision~\cite{he2016deep}.
In those domains it is possible to collect huge amounts of labeled data, the key for deriving robust recognition models that strongly generalize across application boundaries.
In contrast, collecting large-scale, labeled data sets has so far been limited in sensor-based human activity recognition. 
Labeled data in human activity recognition is scarce and hard to come by, as sensor data collection is expensive, and the annotation is time-consuming and sometimes even impossible for privacy or other practical reasons.
A model derived from such a sparse dataset is not likely to generalize well.

Despite the numerous efforts in improving human activity dataset collection, the scale of typical datasets remains small, thereby only covering limited sets of activities~\cite{Chavarriaga13prl,Zhang12ubicomp,thomaz2015practical,hovsepian2015cstress}. 
Even the largest sensor-based activity dataset only spans a few dozen users and relatively short durations~\cite{Bachlin10titb,reiss2012introducing}, which is in stark contrast to the massive datasets in other domains that are often several orders of magnitude larger. 
For example, Daphnet freezing of gait dataset~\cite{Bachlin10titb} has 5 hours of sensor data from 10 subjects, and PAMAP2 dataset~\cite{reiss2012introducing} has 7.5 hours of sensor data from 9 subjects.
However, for reference, the ImageNet dataset~\cite{deng2009imagenet}  has approx.\ 14 million images, and the "One billion words" benchmark ~\cite{chelba2013one} contains literally one billion words.

In this work, {we develop a framework} that can potentially alleviate the sparse data problem in sensor-based human activity recognition.
We aim at harvesting existing video data from large-scale repositories, such as YouTube, and automatically generate data for virtual, body-worn movement sensors (IMUs) that will then be used for deriving sensor-based human activity recognition systems that can be used in real-world settings.
The overarching idea is appealing due to the sheer size of common video repositories and the availability of labels in {the} form of video titles and descriptions. 
Having access to such data repositories opens up possibilities for more robust and potentially more complex activity recognition models that can be employed in entirely new application scenarios, which so far could not have been targeted due to limited robustness of the learned models.
The challenges for making these vast amounts of existing data usable for sensor-based activity recognition are manyfold, though:
\textit{i)} the datasets need to be curated and filtered towards the actual activities of interest;
\textit{ii)} even though video data capture the same information about activities in principle, sophisticated preprocessing is required to match the source and target sensing domains;
\textit{iii)} the opportunistic use of activity videos requires adaptations to account for contextual factors such as multiple scene changes, rapid camera orientation changes (landscape/portrait), the scale of the performer in the far sight, or multiple background people not involved in the activity; and
\textit{iv)} new forms of features and activity recognition models will need to be designed to overcome the short-comings of learning from video-sourced motion information for eventual IMU-based inference.

{Our work is part of a growing number of exciting recent research results that explore the generation of cross-modality sensor data from "data-rich" sources such as video and motion capture in various domains \cite{rey2019let, huang2018deep, xiao2020deep, takeda2018multi}. 
For example, in \cite{huang2018deep} IMU data was synthesized from high-fidelity motion capture data with high temporal and spatial resolutions for computing human pose in real-time. 
On a similar vein, \cite{xiao2020deep, takeda2018multi} generate sensory data from motion capture datasets and demonstrate their effectiveness for activity recognition. 
Most similar to our work, \cite{rey2019let} showed in principle that motion information can be extracted from video and utilized for sensor-based HAR.}

{In this paper, we present a method that allows us to effectively use video data} for training sensor-based activity recognizers, and as such demonstrates the first step towards larger-scale, and more complex deployment scenarios than what is considered the state-of-the-art in the field.
Our approach extracts motion information from arbitrary human activity videos, and is thereby not limited to specific scenes or viewpoints. 
We have developed \textbf{IMUTube}, an automated  processing pipeline 
that:
\textit{i)} applies standard pose tracking and 3D scene understanding techniques to estimate full 3D human motion from a video segment that captures a target activity;
\textit{ii)} translates the visual tracking information into virtual motion sensors (IMU) that are placed on dedicated body positions;
\textit{iii)} adapts the virtual IMU data towards the target domain through distribution matching; and
\textit{iv)} derives activity recognizers from the generated virtual sensor data, potentially enriched with small amounts of real sensor data. 
Our pipeline integrates a number of off-the-shelf 
computer vision and graphics techniques, so that IMUTube is fully automated and thus directly applicable to a rich variety of existing videos.  
One notable limitation from our current prototype is that it still requires human curation of videos to select appropriate activity content. 
However, with advances in computer vision 
the potential of our approach can be further increased towards complete automation.

The work presented in this paper is {our} first step towards the greater vision of automatically deriving robust activity recognition systems for body-worn sensing systems.
The key idea is to opportunistically utilize as much existing data and information as possible thereby not being limited to the particular target sensing modalities.
We present the overall approach and relevant technical details and explore the potential of the approach on practical recognition scenarios.
Through a series of experiments on three benchmark datasets---RealWorld \cite{sztyler2016body}, PAMAP2 \cite{reiss2012introducing}, and Opportunity \cite{Chavarriaga13prl}---we demonstrate the effectiveness of our approach.
We discuss the overall potential of models trained purely on virtual sensor data, which in certain cases can even reach recognition accuracies that are comparable to models that are trained only on actual sensor data. 
Moreover, we show that when adding only small portions of real sensor data during model training we are even able to  outperform those models that were trained on real sensor data alone.
As such, our experiments show the potential of the proposed approach, a paradigm shift for deriving sensor-based human activity recognition systems.

This work opens up the opportunity for the human activity recognition community to expand the general focus towards more complex and more challenging recognition scenarios. 
We expect the proposed approach to dramatically accelerate the progress of human activity recognition research.
With the proposed method it will also be possible to freely experiment with and optimize on-body sensor configurations, which will have a substantial impact on real-world deployments.
We discuss possible extensions to the presented approach, and thus define a research agenda towards next-generation sensor-based human activity recognition. 


\section{EXTRACTING VIRTUAL IMU DATA FROM VIDEOS}
\label{sec:processing-pipeline}

The key idea of our work is to replace the conventional data collection procedure that is typically employed for the development of sensor-based human activity recognition (HAR) systems.
Our approach aims at making existing, large-scale video repositories accessible for the HAR domain, leading to training datasets of sensor data, such as IMUs, that are potentially multiple orders of magnitude larger than what is standard today.
With such a massively increased volume of \textit{real} movement data---in contrast to simulated or generated samples, that often do not exhibit the required quality nor variability---it will become possible to develop substantially more complex and more robust activity recognition systems with a potentially much broader scope than the state-of-the-art in the field.
In what follows, we first give an overview of the general approach before we provide the technical details of our procedure that converts videos into virtual IMU data.

\begin{figure}[t]
\centering
\includegraphics[width=\textwidth]{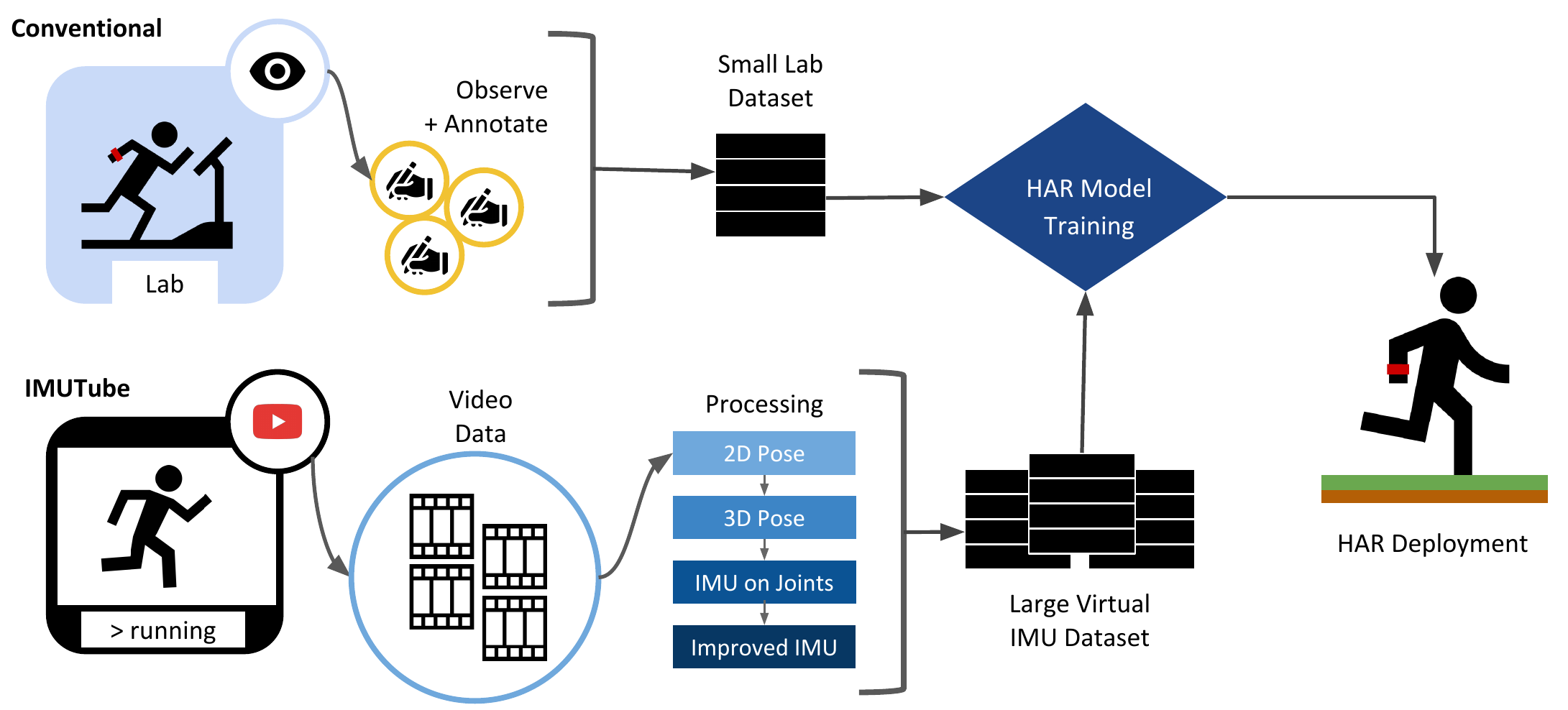}
\caption{The proposed IMUTube system replaces the conventional data recording and annotation protocol (upper left) for developing sensor-based human activity recognition (HAR) systems (upper right). We utilize existing, large-scale video repositories from which we generate virtual IMU data that are then used for training the HAR system (bottom part).}
\label{fig:overall} 
\vspace*{-1em}
\end{figure}

\subsection{IMUTube Overview}
\label{sec:processing-pipeline:overview}
\autoref{fig:overall} gives on overview of {our framework} for deriving sensor-based human activity recognition systems.
The top left part ("conventional") summarizes the currently predominant protocol.
Study participants are recruited and invited for data collection in a laboratory environment.
There they wear sensing platforms, such as a wrist-worn IMU, and engage in the activities of interest, typically in front of a camera.
Human annotators provide ground truth labeling either directly, i.e., while the activities are performed, or based on the video footage from the recording session.
This procedure is very labor-intensive and often error-prone, and, as such, labeled datasets of only limited size can typically be recorded with reasonable efforts.

In contrast, our approach aims at utilizing existing, large-scale repositories of videos that capture activities of interest (bottom left part of \autoref{fig:overall} labeled "IMUTube").
With the explosive growth of social media platforms, a virtually unlimited supply of labeled video is available online that we aim to utilize for training sensor-based HAR systems.
In our envisioned application, a query for a specific activity delivers a (large) set of videos that seemingly capture the target activity.
These results (currently) need to be curated in order to eliminate obvious outliers etc.\ such that the videos are actually relevant to the task (see discussion in Section~\ref{sec:discussion}).
Our processing pipeline then converts the video data into usable virtual sensor (IMU) data.
The procedure is based on a computer vision  pipeline that first extracts 2D pose information, which is then lifted to 3D.
Through tracking individual joints of the extracted 3D poses, we are then able to generate sensor data, such as tri-axial acceleration values, at many locations on the {body}. 
These values are then post-processed to match the target application domain.

{Our} work aims at replacing the data collection phase of HAR development.
It is universal as it does not impose constraints on model training (top center in \autoref{fig:overall}) nor deployment (right part of \autoref{fig:overall}).
In what follows, we describe the technical details of our processing pipeline that make videos usable for training IMU-based activity recognition systems.
This description assumes direct access to a video that captures a target activity, i.e., here we do not focus on the logistics and practicalities of querying video repositories and curating the search results.

\subsection{Motion Estimation for 3D Joints}
\begin{figure}[t]
\begin{center}
\begin{tabular}[t]{@{\hspace{-0.25cm}} c}
    \includegraphics[align=c,width=0.9\linewidth]{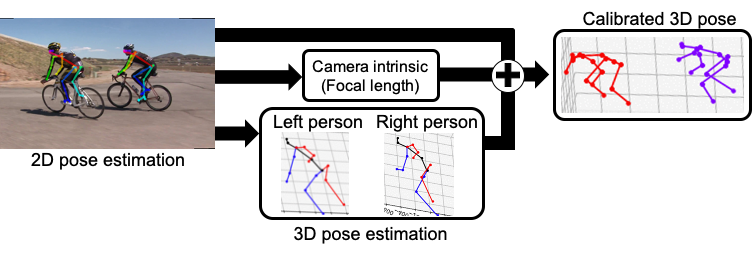}
    \vspace{-0.0in}
\end{tabular}
\end{center}
\vspace{-0.15in}
\caption{3D joint orientation estimation and pose calibration.
The multi-person 2D poses are estimated with \emph{OpenPose} followed by lifting to 3D through \emph{VideoPose3D}.
The camera intrinsic parameters are estimated using the \emph{DeepCalib} model.
Jointly with the pose and camera related parameters, we calibrate the orientation and translation in the 3D scene for each frame.
}
\label{fig:pose} 
\vspace*{-1em}
\end{figure}
On-body movement sensors capture local 3D joint motion, and, as such, our processing pipeline aims at reproducing this information but from 2D video.
As shown in \figref{fig:pose} we employ a two-step approach.
First, we estimate 2D pose skeletons for potentially multiple people in a scene using a state-of-the-art pose extractor, namely the \emph{OpenPose} model \cite{cao2017realtime}.
Then, each 2D pose is lifted to 3D by estimating the depth information that is missing in 2D videos.
Without limiting the general applicability we assume here that all people in a scene are performing the same activity.
Although fast and accurate, the \emph{OpenPose} model estimates 2D poses of people on a frame by frame basis only, i.e., no tracking is included which requires post-processing to establish and maintain person correspondences across frames.
In response, we apply the \emph{SORT} tracking algorithm \cite{Bewley2016_sort} to track each person across the video sequence.
\emph{SORT} utilizes bipartite graph matching with the edge weights as the intersection-over-union (IOU) distance between boundary boxes of people from consecutive frames.
The boundary boxes are derived as tight boxes including the 2D keypoints for each person.

To increase the reliability of the 2D pose detection and tracking, we remove 2D poses where over half of the joints are missing, and also drop sequences that are shorter than one second.
For each sequence of a tracked person, we also interpolate and smooth missing or noisy keypoints in each frame using a Kalman filter, as poses cannot be dramatically different between subsequent frames.
Finally, each 2D pose sequence is lifted to 3D pose by employing the  \emph{VideoPose3D} model~\cite{pavllo20193d}.
Capturing the inherent smooth transition of 2D poses across the {frames} encourages more natural 3D motion in the final estimated (lifted) 3D pose.

\subsection{Global Body Tracking in 3D}

Inertial measurement units capture the acceleration from global body movement in 3D, and additionally local joint {motion} in 3D.
Thus, we also have to extract global 3D scene information from the 2D video to track a person's movement in the whole scene.
Typical 3D pose estimation models do not localize the global 3D position and orientation of the pose in the scene. 
Tracking the {3D} position of {a} person in 2D {video} requires two pieces of information:
\textit{i)} 3D localization in each 2D frame; and 
\textit{ii)} the camera viewpoint changes (ego-motion) between subsequent 3D scenes. We map the 3D pose of a frame to the corresponding position within the whole 3D scene in the video, compensating for the camera viewpoint of the frame. 
The sequence of the location and orientation of 3D pose is the global body movement in the whole 3D space.
For the virtual sensor, the global acceleration from the tracked sequence will be extracted along with local joint acceleration.

\subsubsection{3D Pose Calibration}
First, we estimate the 3D rotation and translation of the 3D pose within a frame, as shown in \figref{fig:pose}.
For each frame, we calibrate each 3D pose from a previously estimated 3D joint according to the perspective projection between corresponding 3D and 2D keypoints.
The perspective projection can be estimated with the {Perspective-n-Point (PnP)} algorithm~\cite{Hesch_2011}.
Additionally to 3D and 2D correspondences, the {PnP} algorithm requires the camera intrinsic parameters for the projection, which include focal length, image center, and the lens distortion parameters~\cite{Shah_1996,Caprile_1990}.
Since arbitrary online videos typically do not come with such metadata, the camera intrinsic parameters are estimated from the video with the \emph{DeepCalib} model~\cite{Bogdan_2018}.
The \emph{DeepCalib} model is a frame-based model that considers a single image at a time so that the estimated intrinsic parameter for each frame slightly differs across the frame according to its scene structure.
Hence, we assume that a given video clip sequence is recorded with a single camera, and aggregate the intrinsic parameter predictions by calculating the average from all frames:
\begin{equation}
    \begin{gathered}
        c^{int} = \frac{1}{T}\sum_{t=1}^T c^{int}_t
    \end{gathered}
\end{equation}
where $c^{int}=[f,p,d]$ is the averaged camera intrinsic parameters from each frame, $x_t$ at time $t$, 
predictions, $c^{int}_t=DeepCalib(x_t)$. $f=[f_x,f_y]$ is the focal length and $p=[p_x,p_y]$ is {the} optical center for {the} $x$ and $y$ axis, and $d$ denotes the lense distortion.
Once the camera intrinsic parameter is calculated, the {PnP} algorithm regresses global pose rotation and translation by minimizing the following objective function: 
\begin{equation}
    \begin{gathered}
        \{ R^{calib}, T^{calib} \} = \arg\min_{R, T} \sum_{i=1}^N \| p^i_2 - \frac{1}{s} c^{int} (R p^i_3 + T) \| \\
        \text{subject to } R^T R = I_3, det(R) = 1
    \end{gathered}
\end{equation}

\noindent where $p_2 \in \mathbb{R}^2$ and $p_3 \in \mathbb{R}^3$ are corresponding 2D and 3D keypoints.
$R^{calib} \in \mathbb{R}^{3 \times 3}$ is the extrinsic rotation matrix, $T^{calib} \in \mathbb{R}^3$ is the extrinsic translation vector, and $s \in \mathbb{R}$ denotes the scaling factor~\cite{Qilong_Zhang,Zhuang_1995}.
For the temporally smooth rotation and translation of a 3D pose across frames, we initialize the extrinsic parameter, $R$, and $T$, with the result from the previous frame.
The 3D pose for each person, $p_3 \in \mathbb{R}^{3 \times N}$, at each frame is calibrated (or localized) with the estimated corresponding extrinsic parameter.
\begin{equation}
    p^{calib}_3 = R^{calib} p_3 + T^{calib}
\end{equation}

From the calibrated 3D poses, $p^{calib}_3 \in \mathbb{R}^{3 \times N}$, we remove people considered as the background.
For example,  
{in a} rope jumping competition scene, a set of people may rope jump while others are sitting and watching.
Depending on the scene, not all people captured may partake in an activity (e.g., bystanders).
To effectively collect 3D pose and motion that belongs to a target activity, we thus remove those people in the (estimated) background.
We first calculate the pose variation across the frames as the summation of the variance of each joint location across time.
Subsequently, we only keep those people with the pose variation larger than the median of all people.

\subsubsection{Estimation of Camera {Ego-motion}}

\begin{figure}[t]
\begin{center}
\begin{tabular}[t]{@{\hspace{0.35cm}} c}
    \includegraphics[align=c,width=0.9\linewidth]{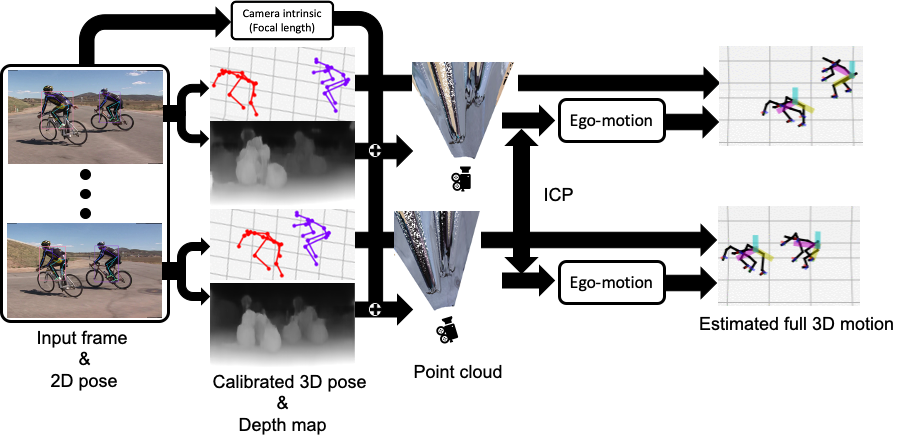}
    \vspace{-0.0in}
\end{tabular}
\end{center}
\vspace{-0.15in}
\caption{3D pose and motion tracking with {compensation of} the camera motion.
The camera motion is estimated through the iterative closest point (ICP) algorithm between subsequent {point clouds}.
Then, calibrated 3D poses per frame are mapped to the location in the entire 3D scene, compensating {for} camera motion.
The calibrated 3D poses from both frames are initially centered in the 3D world origin as the camera follows the cyclists.
After incorporating ego-motion information, we can see that two cyclists are moving from right to left, moving closer to each other as in the video (most right figure).
}
\label{fig:ego_motion} 
\vspace*{-1em}
\end{figure}

In an arbitrary video, the camera can move around the scene freely.
However, the pipeline should not confuse the camera motion with human motion.
For example, a person who does not move (much) may appear at a different location in subsequent frames due to the camera movement, which is misleading for our purpose.
Also, a moving person can always appear in the center of the frame, and thus erroneously appear static, if the camera follows that person and therefore the movements are effectively compensated for in the video.
In these two cases, the virtual sensor should capture no motion (static), or the global body acceleration only, respectively, independently from camera motion.
Hence, before generating the virtual sensor data, the 3D poses, which were previously calibrated per frame, need to be corrected for camera ego-motion, i.e., potential viewpoint changes, across the frames.

Camera ego-motion estimation from one viewpoint to another requires 3D point clouds of both scenes~\cite{Rusinkiewicz,Besl_1992,Pomerleau_2015}.
Deriving a 3D point cloud of a scene requires two pieces of information: 
\textit{i)} the depth map; and 
\textit{ii)} camera intrinsic parameters.
For camera intrinsic {parameters}, we reuse the {parameters} previously estimated.
The depth map is the distances of pixels in the 2D scene from a given camera center, which we estimate with the \emph{DepthWild} model~\cite{Gordon_2019} for each frame.
Once we have obtained the depth map and the camera intrinsic parameters, we can geometrically inverse the mapping of each pixel in the image to the 3D point cloud of the original 3D scene.
With basic trigonometry, the point cloud can be derived from the depth map using the previously estimated camera intrinsic parameter, $c^{int}=[f_x, f_y, p_x, p_y, d]$. 
For a depth value $Z$ at image position $(x,y)$, the point cloud value, $[X, Y, Z]$, is:
\begin{equation}
    [X, Y, Z] = \Big[\frac{(x-p_x) \cdot Z}{f_x}, \frac{(y-p_y) \cdot Z}{f_y}, Z \Big]
\end{equation}

Once point clouds are calculated across frames, we can derive the camera ego-motion (rotation and translation) parameters between two consecutive frame point clouds.
A popular method for registering groups of point clouds is the Iterative Closest Points (ICP) algorithm \cite{Rusinkiewicz,Besl_1992,Pomerleau_2015}. 
Fixing a point cloud as a reference, ICP iteratively finds closest point pairs between two point clouds and {estimates} rotation and translation for the other point cloud that minimizes the positional error between matched points~\cite{Besl_1992}.
Since we extract color point clouds from video frames, we adopted Park \etal's variant of the ICP algorithm \cite{park2017colored}, which considers color matching between matched points in addition to the surface normals to enhance color consistency after registration. 
More specifically, we utilize background point clouds instead of the entire point cloud from a scene because the observational changes for the stationary background objects in the scene are more relevant to the camera movement.
We consider humans in the scene as foreground objects, and remove points that belong to human bounding boxes {determined} from 2D pose detection. 
The reason for this step is that we noticed that including foreground objects, such as humans, leads to the ICP algorithm confusing movements of moving objects, i.e., the humans, and of the camera.
With the background point clouds, we apply the color ICP algorithm between point clouds at time $t-1$ and $t$, $q_{t-1}$ and $q_t$ respectively. 
As such, we iteratively solve:
\begin{equation}
    \{ R^{ego}_t, T^{ego}_t \} = \arg\min_{R, T} \sum_{(q_{t-1}, q_t) \in \mathcal{K}} (1-\delta) \|\mathcal{C}_{q_{t-1}}(f(Rq_t + T)) - \mathcal{C}(q_{t-1})\| + \delta \|(Rq_t + T - q_{t-1}) \cdot n_{q_{t-1}} \|
\end{equation}
where $\mathcal{C}(q)$ is the color of point $q$, $n_q$ is the normal of point $q$. 
$\mathcal{K}$ is the correspondence set between $q_{t-1}$ and $q_t$, and $R^{ego}_t \in \mathbb{R}^{3 \times 3}$ and $T^{ego}_T \in \mathbb{R}^3$ are fitted rotation and translation vectors in the current iteration. 
$\delta \in [0, 1]$ is the weight parameter {that balances the emphasis given to} positional {or} color matches.

The estimated sequence of translation and rotation of a point cloud represents the resulting ego-motion of the camera.
As the last step, we integrate the calibrated 3D pose and ego-motion across the video to fully track 3D human motion.
Previously calibrated 3D pose sequences, $p^{calib}_3$, are rotated and translated according to their ego-motion at frame $t$:
\begin{equation}
    p^{track}_{3_t} = R^{ego}_t p^{calib}_{3_t} + T^{ego}_t
\end{equation}
where $p^{track}_3 \in \mathbb{R}^{T \times N \times 3}$ is the resulting 3D human pose and motion tracked in the scene for the video, $T$ is the number of frames, and $N$ is the number of joint keypoints.
The overall process of compensating camera ego-motion is illustrated in \figref{fig:ego_motion}.

\subsection{Generating Virtual Sensor Data}

Once full 3D motion information has been extracted for each person in a video, we can extract virtual IMU sensor streams from specific body locations.
The estimated 3D motion only tracks the locations of joint keypoints, i.e., those dedicated joints that are part of the 3D skeleton as it has been determined by the pose estimation process.
However, in order to track how a virtual IMU sensor that is attached to such joints rotates while the person is moving, we also need to track the orientation change of that local joint.
This tracking needs to be done from the perspective of the body coordinates.
The local joint orientation changes can be calculated through forward kinematics based from the hip, i.e., the body center, to each joint.
We utilize state-of-the-art 3D animation software {-- \emph{Blender}~\cite{blender}}, to estimate and track these orientation changes.
Using the orientation derived from forward kinematics, the acceleration of joint movements in the world coordinate system is then transformed into the local sensor coordinate system.
The angular velocity of the {virtual} sensor ({i.e., a} gyroscope) is calculated by tracking orientation changes.

We employ our video processing pipeline on raw 2D videos that can readily be retrieved {by}, for example, querying public repositories such as YouTube, and {combined with} subsequent curation (not within the focus of this paper). 
The pipeline produces virtual IMU, for example, tri-axial accelerometer data.
This data effectively captures the recorded activities, yet the characteristics of the generated sensor data {will still} differ from real IMU data{, for instance it will lack any MEMS noise}. 
In order to compensate for this mismatch, we employ the {\emph{IMUSim}}~\cite{YoungLA11} {model to extract} realistic sensor behavior for each on-body location. 
{\emph{IMUSim}} estimates sensor output considering mechanical and electronic components in the device, as well as the changes of a simulated magnetic field in the environment. 
As such, this post-processing step leads to more realistic IMU data \cite{Phm2018SplineFS,Karlsson14inertial,bodynets}.

\subsection{Distribution Mapping for Virtual Sensor Data}
\label{dist.mapping}
{As the last step before using the virtual IMU dataset for HAR model training, we define a calibration operation to account for any potential mismatch between the source (virtual) and target (real) domains~\cite{10.1145/3380985}. We employ a distribution mapping technique to fix such mismatch, where }we transfer the distribution of the virtual sensor to that of the target sensor. For computational efficiency, the rank transformation approach~\cite{conover1981rank} is utilized:


\begin{equation}
    x_r = G^{-1}(F(X \leq x_{v}))
\end{equation}
where, $G(X\leq x_r) = \int_{-\inf}^{x_r} g(x) dx$ and $F(X\leq x_v) = \int_{-\infty}^{x_v} f(x) dx$ are cumulative density functions for real, $x_r$, and virtual, $x_v$, sensor samples, respectively.
In our experiments {(Section ~\ref{sec:exp2_domainshift})}, we show that only a few seconds to minutes of real sensor data 
is sufficient to {calibrate} 
the virtual sensor effectively for successful activity recognition.

\section{TRAINING ACTIVITY RECOGNITION CLASSIFIERS WITH VIRTUAL IMU DATA} \label{sec:exp1}


We now describe a series of experiments to examine the viability of using IMUTube to produce virtual IMU data useful for HAR. 
Our first set of experiments {consider} the performance of virtual IMU data on a HAR dataset providing both video and real IMU data, which enables a fair comparison between virtual and real IMU data. Here, we see promising results suggesting that training activity classifiers from virtual IMU data alone can perform well on real IMU data. 
We then move on to show that activity classifiers trained using this virtual IMU data can also perform well on real IMU data coming from common HAR datasets, namely Opportunity \cite{Chavarriaga13prl} and PAMAP2 \cite{reiss2012introducing}. 
Finally, we describe how we curate a video dataset comprising of online videos (e.g.,  YouTube) in order to extract virtual IMU data for complex activities.

In each experiment, we compare the performances of models on real IMU data (i.e., the test data is from real IMUs), when trained from real IMUs (R2R), trained from virtual IMUs (V2R), or trained from a mixture of virtual and real (Mix2R) IMU data. 
{Throughout our experiments, we consider the Random Forest classifier as our main machine learning back-end for activity recognition, evaluated via leave-one-subject-out scheme. We supplement this primary result by also demonstrating the feasibility to apply deep learning with a hold-out evaluation scheme; in doing so we show our approach is agnostic to the choice of the learning algorithm.}




\subsection{Feasibility Experiment under Controlled Conditions} \label{sec:exp1_rw}
There are many potential sources of noise which may impact the activity recognition performance; therefore, in our first experiment we hold constant as many factors as possible. We accomplish this by using the RealWorld dataset~\cite{sztyler2016body}, an activity recognition dataset that contains not only IMU data but also provides videos of the subjects performing the activities. 

\subsubsection*{Data} \, The Realworld dataset covers $15$ subjects performing eight locomotion-style activities, namely \emph{climbing up, climbing down, jumping, lying, running, sitting, standing}, and \emph{walking}. Each subject wears the sensors for approximately ten minutes for each activity except for jumping (<2 minutes). The video and accelerometer data are not time-synchronized, as each video starts some time (under one minute) before each activity begins. The video is recorded using a hand-held device by the experiment's {administrator}, who follows the subject as they perform the activity (e.g., running through the city alongside the subject). The videos do not always present a full-body view of the subject, and the video-taker sometimes makes arbitrary changes to the video scene (e.g., he/she might walk past the subject, or rotate the camera from landscape to {portrait} mode halfway). These factors present extra difficulty in extracting virtual IMU for the full duration of the activities; nonetheless we are able to extract 12 hours of virtual IMU data, this is compared to 20 hours of available real IMU data. As a preprocessing step, we remove the first ten seconds of each video and divide the remainder into two-minute chunks for efficient running of IMUTube. {Virtual IMU data are extracted from 7 body locations, i.e. forearm, head, shin, thigh, upper arm and waist/chest, corresponding to where real sensors are placed in Realworld.} We assume all IMU data to have a frequency of 30Hz and use sliding windows to generate training samples of duration 1 second and 50\% overlap. The resulting real and virtual IMU dataset contains $221k$ and $86k$ windows,  respectively.



\subsubsection*{Method} \, {Our primary evaluations are performed with the Random Forest classifier using ECDF features \cite{hammerla2013preserving} ($15$ components), trained using a leave-one-subject-out scheme. On Realworld, we use a train set of 13 subjects, validation set of 1 subject and test set of 1 subject in each fold. This scheme is followed in R2R (where training data is real IMU data), V2R (where training data is virtual, and distribution-mapped only using data from train users), and Mix2R (which contains a mixture of both real and virtual IMU data). We calibrate hyperparameters on the validation subjects by varying the number of trees from 3 to 50 and the minimum number of samples in leaf node from 1 to 50. We report the mean F1-score of the test subjects computed after the completion of all folds.} 

{Separate from this, we train DeepConvLSTM~\cite{Ordonez16sensors} on a hold-out evaluation scheme, where subject 15 is randomly selected as validation, 14 as test, and the rest as the training set.} DeepConvLSTM is trained on raw data for a maximum of 100 epochs with an Adam optimizer \citep{kingma2014adam} and early stopping on the validation set with a patience of ten epochs; learning rate is searched from $10^{-6}$ to $10^{-3}$, and weight decay from $10^{-4}$ to $10^{-3}$ via grid search. We further regularize model training by employing augmentation techniques from \cite{um2017data} with a probability of application set at either $0$ and $0.5$ depending on validation set result. We average over 3 runs initiated with different random seeds and report the mean F1-score.

In both cases, we report the highest test F1-score achieved using any amount of training data, along with the Wilson score interval with 95\% confidence. 
We discuss {the} effect of training {set size} in Section \ref{sec:exp2_mix2r}. 
We reuse these settings throughout the paper unless stated otherwise.

\subsubsection*{Results} \, {In \autoref{tbl:rw_rf}, we see convincing evidence that human activity classifiers can learn from virtual IMU data alone. 
When learning from virtual IMU data alone (V2R), the {8-class} model achieves an F1-score of 0.57, which is within 2\% of that achieved by learning from real IMU data (R2R). This result is remarkable as the difference in recognition performance of R2R and V2R is small notwithstanding the change in data source and the introduction of noise while going through our pipeline.}

{Furthermore, when we use a mixture of virtual and real IMU data to train the model, it is even able to surpass R2R performance with a significant relative performance gain of 12\%, reaching an F1-score of 0.64. This showcases an additional potential of IMUTube -- we can build activity classifiers using both virtual and real IMU data to push recognition capabilities beyond that achieved by either.}

{Our DeepConvLSTM results (evaluated on a random subject, \autoref{tbl:rw_lstm}) offers another perspective into modeling virtual IMU data when deep learning models are used.} Although learning from virtual IMU data alone is seen to pose more challenges (V2R achieves 75\% of R2R), this is possibly related to the setup of learning directly from raw data, in contrast to processed features in the Random Forest case. As a consequence, DeepConvLSTM may be learning feature representations highly specific to the virtual IMU domain, which prevents immediate generalization to real IMU data. This issue is {diminished} when using a mixture of virtual and real IMU data for training, as Mix2R even outperforms R2R by 6.6\%. We presume that the improvement is related to the complementary benefits of both real and virtual data, as well as the feature learning capabilities of deep learning models, which learn better when more data is available. 

This set of results provide promising signs {for} IMUTube -- we can learn capable activity classifiers with virtual IMU data alone, despite having only so far considered relatively straightforward techniques in extracting and modeling the virtual IMU data. We delve into these concerns about the quality of virtual IMU data in Section~\ref{sec:exp2} to provide a more complete view.

\begin{table}[t!]
    \caption{\small Recognition results on the Realworld dataset (8 classes) when training models from real IMU data (R2R), from virtual IMU data (V2R), and from a mixture of both (Mix2R). Wilson score confidence intervals are shown. For Random Forest models, V2R achieves {98\%} of the R2R F1-score, while a Mix2R setup surpasses R2R by {12\%.}
    }
    \label{tbl:rw}
    \begin{subtable}[t]{.5\textwidth}
        \caption{\small {Random Forest (leave-one-subject-out)}}
        \label{tbl:rw_rf}
        \raggedright
            \begin{tabular}{c|c|c}
                \hline R2R  & V2R & Mix2R \\
                \hline \hline
                0.5779$\pm0.0025$ & 0.5675$\pm0.0025$ & 0.6444$\pm0.0024$ \\
                \hline\hline
            \end{tabular}
    \end{subtable}%
   \begin{subtable}[t]{.5\textwidth}
        \raggedleft
        \caption{\small {DeepConvLSTM (random single-subject hold-out)}}
        \label{tbl:rw_lstm}
        \begin{tabular}{c|c|c}
            \hline
                R2R  & V2R & Mix2R \\
                \hline \hline
                0.7305$\pm0.0073$ & 0.5465$\pm0.0082$ & 0.7785$\pm0.0068$ \\
        \hline\hline
        \end{tabular}
    \end{subtable}
\end{table}

	

\subsection{Performance on Common Activity Recognition Datasets}

\begin{table}[t!]
    \centering
    \caption{\small Recognition results (mean F1-score) on Opportunity dataset (4 classes) and locomotion activities found in PAMAP2 (8 classes) when using different training data. For Random Forest models, V2R achieves {95\%} of R2R F1-scores on average, while Mix2R outperforms R2R by {5\%} on average.}
    \label{tbl:opp_p8}
    \begin{subtable}[b]{.55\textwidth}
        \caption{\small {Random Forest (leave-one-subject-out)}}
        \label{tbl:opp_p8_rf}
        \raggedright
            \begin{tabular}{c||c|c|c}
                \hline Dataset  & R2R & V2R & Mix2R \\
                \hline \hline
                Opportunity & 0.8271$\pm0.0034$ & 0.7757$\pm0.0037$ & 0.8820$\pm0.0029$ \\
                PAMAP2 (8-class) & 0.7029$\pm0.0055$ & 0.6728$\pm0.0058$ & 0.7284$\pm0.0053$ \\
                \hline\hline
            \end{tabular}
    \end{subtable}%
    \bigskip 
    
   \begin{subtable}[b]{.55\textwidth}
        \raggedleft
        \caption{\small {DeepConvLSTM (random single-subject hold-out)}}
        \label{tbl:opp_p8_lstm}
        \begin{tabular}{c||c|c|c}
            \hline
                Dataset  & R2R & V2R & Mix2R \\
                \hline \hline
                Opportunity &0.8871$\pm0.0074$ &0.7882$\pm0.0096$ & 0.8838$\pm0.0075$ \\
                PAMAP2 (8-class) & 0.7002$\pm0.0161$ &0.5524$\pm0.0175$ & 0.7020$\pm0.0161$ \\
        \hline\hline
        \end{tabular}
    \end{subtable}
\end{table}

We have achieved promising results under the controlled conditions of Realworld, which simultaneously gathers video and IMU together. 
We now seek to relax these conditions, and establish the viability of IMUTube when the exact actions performed in the video data and the real IMU data do not completely align. 
Imagine a scenario where we want to build a classifier for `standing' vs. `sitting'.
Instead of collecting simultaneous video and real IMU data of people standing and sitting, we want to leverage existing videos of people standing and sitting and train the classifier using the derived virtual data.

In the following, we test this scenario by re-using the video data from Realworld and learning models from its virtual data to test on two common HAR datasets, Opportunity and PAMAP2. These datasets are considered as they contain activity labels that roughly correspond to those in Realworld. 




\subsubsection*{Data} \, We consider activities in Opportunity and PAMAP2 which are overlapping with those in Realworld, i.e., four classes (\emph{stand, walk, sit, lie}) in Opportunity, and eight classes (\emph{ascending stairs, descending stairs, rope jumping, lying, running, sitting, standing, {walking}}) in PAMAP2. We use 1-second sliding windows with $50\%$ overlap. 

For Opportunity, we re-extracted virtual data from the Realworld videos in eleven body positions (left and right feet, left shin and thigh, hip, back, left and right arms, left and right forearms), which resulted in $40k$ and $46k$ real and virtual IMU windows respectively. {For DeepConvLSTM, we used random subject 3 for validation, 4 for test, and the rest for training.}

For PAMAP2, the activities are slightly different from those in Realworld so we equated the labels with the closest meaning (e.g., using \emph{jumping} Realworld videos as the source for \emph{rope jumping} virtual IMU in PAMAP2. The PAMAP2 dataset specifies that sensors were placed in three locations (dominant wrist, dominant ankle, chest), which gives rise to a total of four possible combinations for arm and chest location when we extract virtual IMU data from a single video (i.e., left-left, right-right, left-right, right-left). We took advantage of this ambiguity and extracted $4\times$ as much virtual IMU per video, resulting in $24k$ and $152k$ windows for real and virtual IMU respectively. {For DeepConvLSTM, we followed the same setup as \cite{Hammerla16ijcai} and use subject 5 for validation, 6 for test, and the rest for training.}

\subsubsection*{Results} \, \autoref{tbl:opp_p8_rf} shows the classification performance for R2R, V2R and Mix2R. {We observe encouraging results, where learning from virtual IMU data alone can recover high levels of R2R performance, despite data collection conditions not being held constant. A Random Forest classifier trained from virtual IMU data achieves 94\% and 96\% of R2R performance on Opportunity and PAMAP2 respectively.} While this good performance might be related to the simplicity of the motions classified ({mainly locomotive activities}), we highlight that the conditions of data collection in Realworld and Opportunity are very different--subjects could be walking through the forest or city in Realworld, but all subjects perform activities inside a laboratory in Opportunity; Likewise for Realworld and PAMAP2--subjects could also be climbing down the streets of a city (a mixture of pavement and stairs) in Realworld whereas all subjects are climbing up the same building in PAMAP2. Thus, being able to utilize virtual data from one scenario and test it on another is not a trivial task. These results suggest that, on these two tasks, virtual IMU data can provide salient features that are generalizable and robust across testing scenarios.

{We also observe performance gains when training with a mixture of real and virtual IMU data, which exceeds R2R F1-scores by 5\% on average. Not only does this observation solidify the argument that virtual and real IMU data can bring complementary benefits to activity recognition, but such performance gains are also a positive sign especially since the two types of data are collected under rather different circumstances. We argue that, adding virtual IMU data -- in this case, virtual data generated from a related different scenario -- can help expand the variety of motions seen by the classifier and as a result improve model generalization.}

{As before, we provide the performance by DeepConvLSTM on a random test subject as additional results in \autoref{tbl:opp_p8_lstm}, where V2R recovers 84\% of R2R F1-scores, while Mix2R and R2R scores are statistically comparable.}

Overall, this set of results presents strong evidence supporting the usefulness of virtual IMU data, either used standalone or in combination with real IMU data for activity recognition. Beyond this, these results also imply an encouraging view that aligns well with our vision for IMUTube -- that virtual data, even when collected under vastly different settings, can be useful in building capable or even better models for activity recognition.

\subsection{Virtual IMU Data for Complex Activity Recognition}

\begin{table}[t!]
    \caption{\small Recognition results (mean F1-score) on  PAMAP2 (11-classes) when using different training data. The Random Forest model trained with virtual IMU data including YouTube videos (for four complex activities) recovered {80\%} of R2R model performance. 
	For Mix2R, additional real IMU data helped the Random Forest model increase performance up to 98\% of R2R model performance.}
    \label{tbl:p11}
    \begin{subtable}[t]{.5\textwidth}
        \caption{\small {Random Forest (leave-one-subject-out)}}
        \label{tbl:p11_rf}
        \raggedright
            \begin{tabular}{c|c|c}
                \hline R2R  & V2R & Mix2R \\
                \hline \hline
                0.7225$\pm0.0044$ & 0.5792$\pm0.0049$ & 0.7111$\pm0.0044$ \\
                \hline\hline
            \end{tabular}
    \end{subtable}%
   \begin{subtable}[t]{.5\textwidth}
        \raggedleft
        \caption{\small {DeepConvLSTM (random single-subject hold-out)}}
        \label{tbl:p11_lstm}
        \begin{tabular}{c|c|c}
            \hline
                R2R  & V2R & Mix2R \\
                \hline \hline
                0.6977$\pm0.0129$ &0.5326$\pm0.0140$ & 0.7095$\pm0.0128$ \\
        \hline\hline
        \end{tabular}
    \end{subtable}
\end{table}

Encouraged by the results so far, we now try to apply IMUTube onto activity recognition scenarios with even more challenging conditions and test its ability in building classifiers for complex activities.
Our ultimate vision for IMUTube is to extract virtual data from any video, especially those freely available in large online repositories such as YouTube. To test the feasibility of doing so, we first need to curate a dataset with activity videos {originating} from the web. In the following, we {attempt} to source these videos for complex activities present in PAMAP2, and {train} classifiers with the extracted virtual IMU data. 


\subsubsection*{Data} \, We curated a dataset of virtual data covering four complex activities present in PAMAP2, namely \emph{vacuum cleaning, ironing, rope jumping} and \emph{cycling}. To efficiently locate such videos, we extract annotated video segments from activity video datasets in the computer vision domain, including ActivityNet~\cite{caba2015activitynet}, Kinetics700~\cite{carreira2019short}, HMDB51~\cite{kuehne2011hmdb}, MPIIHPD~\cite{andriluka14cvpr}, UCF101~\cite{soomro2012ucf101}, Charades~\cite{sigurdsson2016hollywood}, AVA~\cite{gu2018ava}, MSRdailyactivity3D~\cite{5543273}, and NTU RGB-D~\cite{Liu_2019_NTURGBD120}. 
The resulting video dataset consists of a mix of videos collected in experiment scenarios and in-the-wild (e.g., from YouTube). In total, we collected $\sim{10}$ hours of virtual data from 7,255 videos. To extend our activity recognition task to as many classes in PAMAP2 as possible, we also reuse the other seven videos from Realworld (we do not use the \emph{jumping} videos); this allows us to consider an 11-class activity recognition problem in PAMAP2. 
As mentioned for the PAMAP2 (8-class) task, we face an ambiguity in sensor location which led us to extract $4\times$ virtual data per video. Using sliding windows of 1-second size and $50\%$ overlap, resulted in $38k$ real and $390k$ virtual IMU windows in total.

\subsubsection*{Results} \, {For these challenging conditions (using in-the-wild videos, learning complex activities), \autoref{tbl:p11_rf} shows that virtual IMU data can still be useful for training activity classifiers. 
With the Random Forest classifier, training from virtual IMU data alone achieves a 0.58 F1-score under V2R, which is 80\% of that achieved with R2R (0.72 F1-score). 
This is a weaker result compared to those achieved on previous datasets (where V2R achieved 96\% of R2R on average).
However, this is because there is an even more drastic difference in the data sources and activity label interpretations between the real and virtual IMU data. 
Another likely factor causing the weaker performance is the quality of virtual IMU data that IMUTube is currently able to produce, which might be amplified by the complex activities introduced in this experiment.
In Section~\ref{sec:exp2_compare} we will examine the fidelity of virtual IMU data, and provide directions to improve its quality in Section~\ref{sec:limitations}. Finally, one must consider as a factor the greater domain shift that is probably present between train and test scenarios, which we will discuss in Section~\ref{sec:exp2_domainshift}.}

{When real IMU data is added to virtual IMU data for training, the Random Forest model gains 23\% and achieves a F1-score of 0.71 in Mix2R versus 0.58 in V2R, though Mix2R is still 2\% worse than R2R performance. We believe these results are related to the domain shift within the training data.} To better cope with the scenario where we want to make use of both real and virtual IMU data, we investigate the effect of mixing data in Section \ref{sec:exp2_mix2r} and investigate more sophisticated methods beyond simple mixing in Section~\ref{sec:exp3}. {Our evaluation under DeepConvLSTM is shown in Table~\ref{tbl:p11_lstm}, and results align with those of the Random Forest.}

Through this set of experiments, we have demonstrated that, despite very challenging conditions--in-the-wild videos and complex activity recognition, it is still feasible to learn capable classifiers using virtual IMU data. 
{The results overwhelmingly support that virtual IMU data generated via IMUTube are useful for even real-world instances of activity recognition. Demonstrated over a range of locomotion and more complex activities, virtual IMU data is seen to effectively capture motion information, such that classifiers can be trained from them alone and still perform well on real IMU data. In addition, mixing real and virtual IMU data for training is also shown to be a potential source of performance gain.}

\section{UNDERSTANDING VIRTUAL IMU DATA} \label{sec:exp2}
Across multiple datasets, the model trained on the virtual IMU dataset (V2R) performed well {on} the real IMU test datasets. The V2R performance varies between {80\% - 90\%} compared to R2R models, and only {matches or outperforms} R2R when trained alongside real IMU data (Mix2R). 
{Although notably}, for the experiment on PAMAP2 (11-class), the V2R model could not outperform R2R even when trained with the larger virtual dataset extracted from multiple video sources.
Thus, in this section, we investigate the potential sources of such performance gaps in detail. First, we analyze the extracted virtual IMU data by inspecting the sample-level similarity in IMU signals using synced sequences of real and virtual IMU data. Then, at a distribution level, we investigate the effects of domain shift, along with the impact of our distribution mapping {technique} (Section \ref{dist.mapping}). Finally, we investigate the mixing of real and virtual IMU data for model training (Mix2R), which was seen to give comparable, if not superior, performance relative to R2R in Section~\ref{sec:exp1}. Through our analysis, we aim to provide key insights into the IMUTube pipeline and the use of virtual IMU data for human activity recognition. {All experiments presented in this section are carried out using Random Forest in the leave-on-subject-out setting unless otherwise specified.}



\begin{figure}[t]
\footnotesize
\begin{center}
\begin{tabular}[t]{c}
(a) Left wrist \\
 \includegraphics[align=c,height=0.3\linewidth]{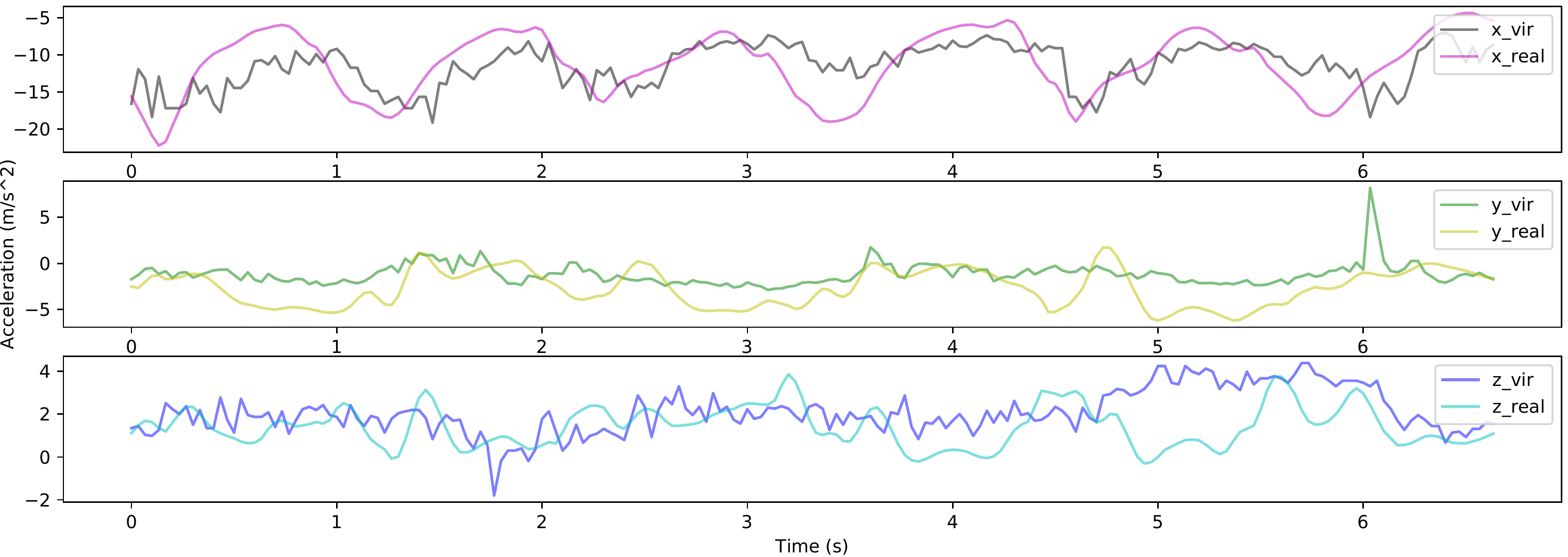}\vspace{-0.0in} \\
    (b) Left ankle \\
    \includegraphics[align=c,height=0.3\linewidth]{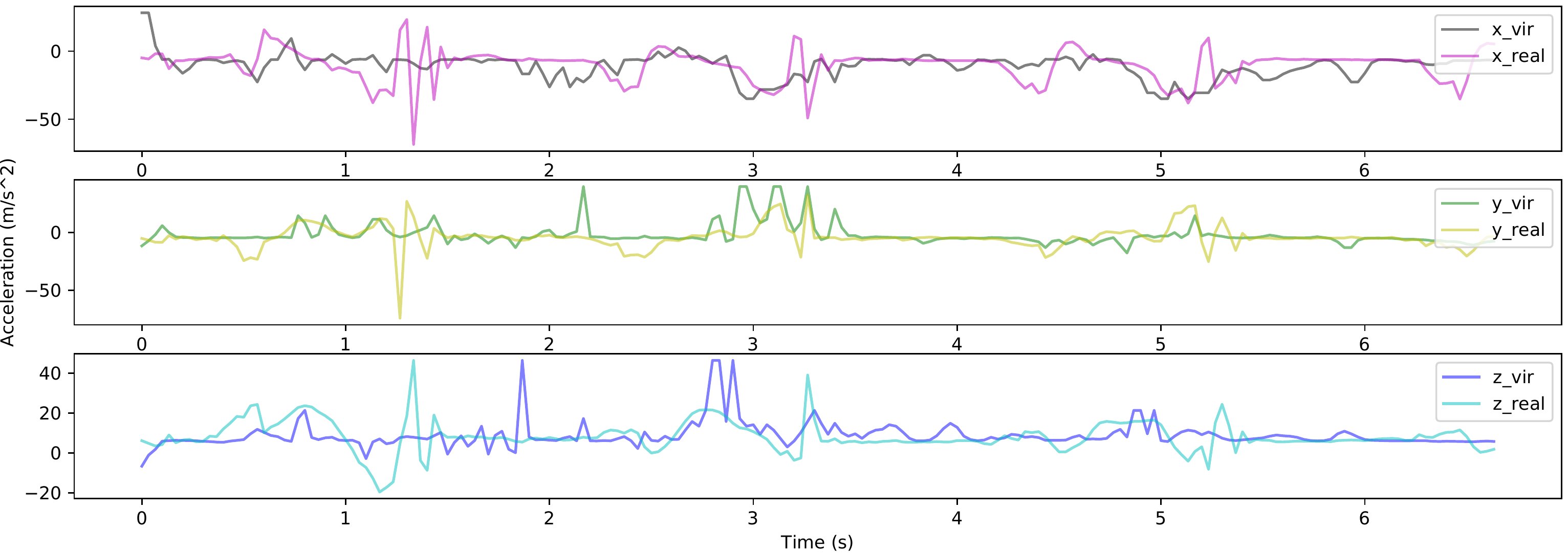}\vspace{-0.1in} \\
\end{tabular}
\end{center}
\caption{Comparison between virtual and real IMU on the TotalCapture dataset. Distribution mapping has been applied to the virtual IMU data. 
}
\label{fig:compare} 
\end{figure}

\subsection{Comparing Virtual and Real IMU} \label{sec:exp2_compare}
{We do not expect IMUTube to function flawlessly.} Given the complexity of the process, the translation from video to virtual IMU data will naturally contain errors.
%
Despite this, we observe promising results of competitive V2R performance in the prior section.
This seems to suggest that perfect sample-level realism in the virtual IMU data is not necessary to train capable human activity classifiers. 
In the following, we compare virtual and real IMU samples to better understand the limits of IMUTube and argue that, the focus, during virtual IMU generation, should be placed on capturing salient features useful for activity recognition.

\subsubsection*{Method} \, 
Sample-level comparison between the virtual and real IMU data requires a dataset with time-synchronized video and IMU sensor {data}.
Although the Realworld dataset contains both accelerometer and video data, these modalities are not synchronized (as mentioned in Section~\ref{sec:exp1_rw}).
Therefore, in this experiment, we introduce the TotalCapture dataset \cite{Trumble:BMVC:2017} which contains time-synchronized (real) IMU data and video recordings (from which the virtual IMU data are extracted). As TotalCapture contains various scripted motions but not labels that are immediately useful for activity recognition-related tasks, we did not evaluate this dataset in Section~\ref{sec:exp1}.

\subsubsection*{Analysis} \, \autoref{fig:compare} shows an example of the virtual and real IMU time-series data of a subject walking, with sensors placed on their wrist and ankle. Along the x-axis, virtual IMU readings are seen to reflect large movement changes {also observed} in the real IMU---one can almost see from the `wrist' time-series (see Figure \ref{fig:compare}(a)) that the person is walking with periodic hand movements. Along the z-axis, the virtual IMU is also seen to capture any spikes in acceleration reasonably well, albeit with a noticeable time lag in the `ankle' case. Virtual and real IMU data differ the most along the y-axis. We postulate that this is related to a dimensionality issue---we are trying to reconstruct 3-D information from a 2-D image time series.
The y-axis here refers to the axis pointing perpendicular to the {visual plane}, which means any acceleration measured along this dimension cannot be easily deduced visually. 

While generating realistic virtual IMU data is important, it is secondary to our main goal of producing virtual IMU data that captures useful information for HAR tasks. To achieve this, what is vital is the ability of the virtual IMU data to capture salient features of the activities that we need to recognize. We already see signs of this happening with the current IMUTube (e.g., the x-axis of the `arm' while walking in \autoref{fig:compare}). This also offers a possible explanation for the better V2R performance seen in predicting locomotion-style activities in Section~\ref{sec:exp1}. Perhaps IMUTube, in its current form, is best suited to capture information about simple motions (i.e., ones mostly characterized by movement in a 2D plane) of which there is still a wide variety, and to which existing HAR methods still struggle to generalize \cite{10.1145/2030112.2030160} (for additional qualitative observations see Section ~\ref{sec:limitations}). To apply IMUTube to more complex activities, it may require improved techniques during virtual IMU data generation (also discussed further in Section~\ref{sec:limitations}).

\subsection{Coping with Domain Shift} \label{sec:exp2_domainshift}

The last step of our pipeline performs a distribution mapping post-processing step between virtual and real data (Section \ref{dist.mapping}). Applying some form of distribution mapping is necessary due to the presence of \emph{domain shift} between training and testing data. This domain shift is not exclusive to extracting virtual sensor data from videos, but it is also present whenever data is taken from different tasks (or datasets) which result in dissimilar data distributions between training and testing~\cite{10.1145/3380985}. 

\subsubsection*{Method} \, 
{Our first experiment is to compare the recognition performance on the Opportunity dataset with models trained using data from sources other than Opportunity, with or without distribution mapping. Specifically, the train data can be \textit{i)} real IMU data from PAMAP2, \textit{ii)} real IMU data from Realworld, or \textit{iii)} virtual IMU data from Realworld videos (Section~\ref{sec:exp1_rw}). Without distribution mapping, we use all available data in the respective datasets that fall under the 4 Opportunity classes (stand, walk, sit, lie) for training, and test using the entire Opportunity dataset. With distribution mapping, we follow a leave-one-subject-out evaluation scheme; In each fold, we train the model using data distribution-mapped with data only from the corresponding train subjects in Opportunity, and evaluate on the remaining test subject.}

{In our second experiment, we focus on the virtual and real IMU used in the 4 datasets described in Section~\ref{sec:exp1}, i.e., Realworld, Opportunity, PAMAP2 8-class and 11-class. We aim to understand how much real IMU data is needed for distribution mapping on the virtual IMU data. To do so, we vary the amount of real IMU data used for distribution mapping and evaluate at what point do the virtual and real IMU data distribution become sufficiently similar. We report the similarity between each data distribution using the Frechet Inception Distance (FID), a metric commonly used in generative modeling to compare the real and generated datasets~\cite{lucic2018gans,heusel2017gans}; lower FID scores indicate more similar data distributions. We also report the confidence interval as calculated by randomly sampling real IMU data with 10 different random seeds for distribution mapping. 
}

\begin{table}[t!]
	\centering
		\caption{\small {Recognition results (mean F1-score, Random Forest) on the 4-class activity recognition task from Opportunity when using training data from other data sources (top rows) and from Opportunity itself (last row, provided for reference). Without distribution mapping, there is a significant drop in performance when using training data collected under different circumstances than the test case, regardless of whether the IMU data is real (66\%) or virtual (64\%). This is resolved when distribution mapping is applied.} } \label{tbl:domainshift}
	\begin{tabular}{c||c||c}
		\hline
		Train data source & Without mapping & With mapping \\
		\hline\hline
		Virtual data & 0.2949$\pm0.0041$ & 0.7757$\pm0.0037$ \\
		PAMAP2       & 0.2770$\pm0.0040$ & 0.6931$\pm0.0041$ \\
		Realworld    & 0.2828$\pm0.0041$ & 0.6637$\pm0.0043$ \\
		\hline\hline
		Opportunity  & 0.8272$\pm0.0034$ & - \\
		\hline\hline
	\end{tabular}
\end{table}

\begin{figure}[t]
\centering
\vspace*{-1em}
\includegraphics[width=\textwidth]{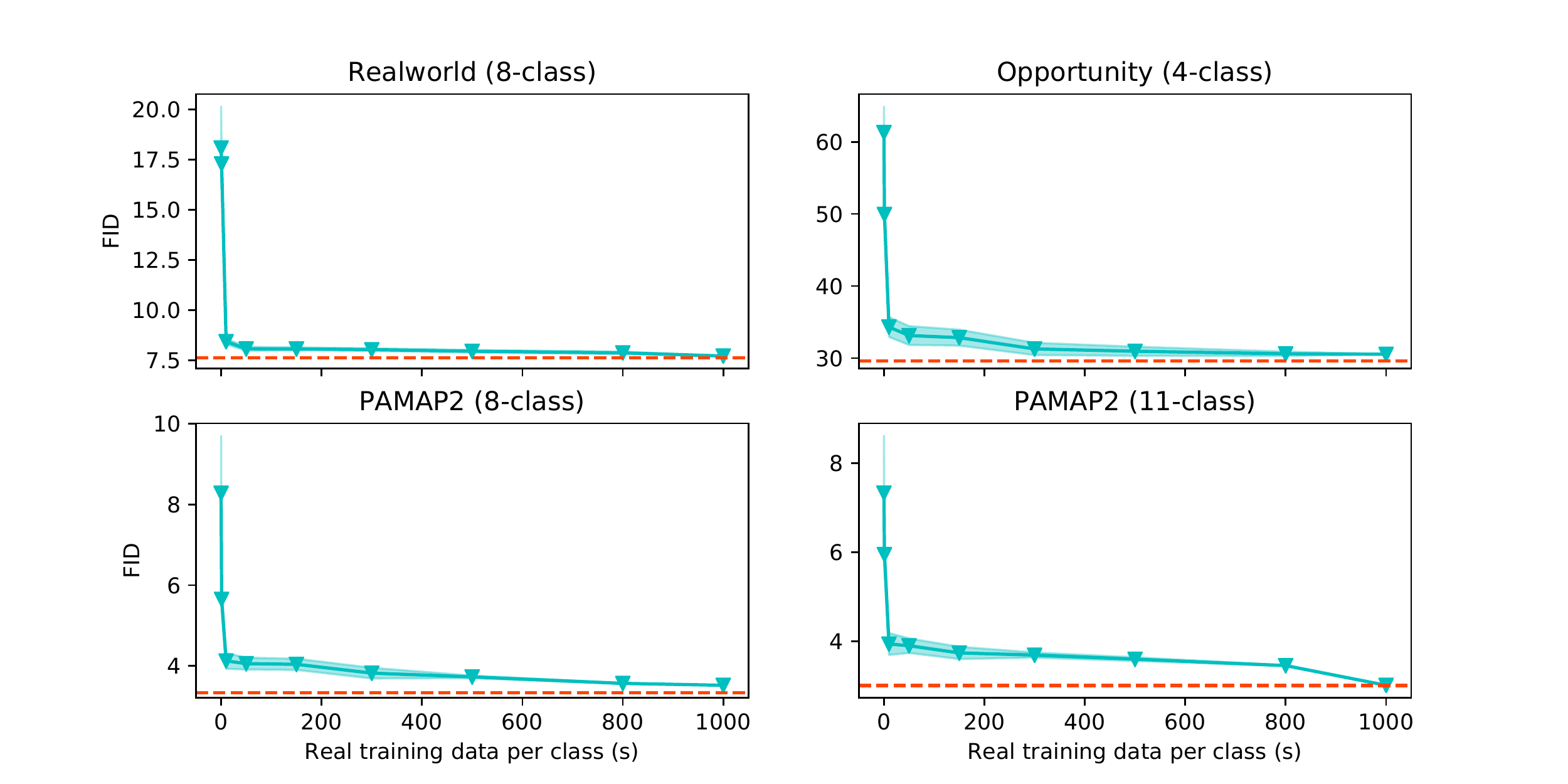}
\caption{
\small {Frechet Inception Distance (FID) and confidence interval (shaded area) between virtual IMU and real IMU data distributions after performing distribution mapping of the virtual IMU data with increasing amounts of real IMU samples. The dotted horizontal line is the FID score obtained when the entire real IMU dataset is used for distribution mapping.}
}  
\label{fig:distmap}
\vspace*{-1em}
\end{figure}

\subsubsection*{Analysis} \, 
{In \autoref{tbl:domainshift}, the effects of domain shift are demonstrated by the significant drop in the performance seen in the `without mapping' column. When using training data not from Opportunity -- despite having the same activity labels -- even models trained with real IMU data (from PAMAP2 and Realworld) suffer a 66\% drop in F1-scores. From this, it is clear that the domain shift issue is not exclusive to the shift present between virtual and real IMU domains.} The drop in performance is resolved when we perform distribution mapping (Section \ref{dist.mapping}); we even observe that training from virtual data outperforms training using other real IMU datasets. This hints that virtual data might have greater value than real IMU data in developing general HAR models. This conclusion was also supported by the results seen when performing the same analysis on the PAMAP2 and Realworld datasets.


{
\autoref{fig:distmap} shows how the virtual/real FID score varies with the quantity of real IMU data used for distribution mapping. In all four cases, an abrupt, significant drop in the FID score is seen with the use of under 100 seconds of real IMU samples. When using $10$ minutes of real IMU data per class for distribution mapping, the FID scores are within 6\% of the final FID score (when all real IMU is used for distribution mapping). 
}

\edit{
}

\subsection{Varying the Mixture and Size of Training Data} \label{sec:exp2_mix2r}
Here, we inspect how varying the mixture and size of training data affects recognition performance on the four datasets considered in Section~\ref{sec:exp1}.


\begin{figure}[t]
\centering
\vspace*{-0em}
\includegraphics[width=\textwidth]{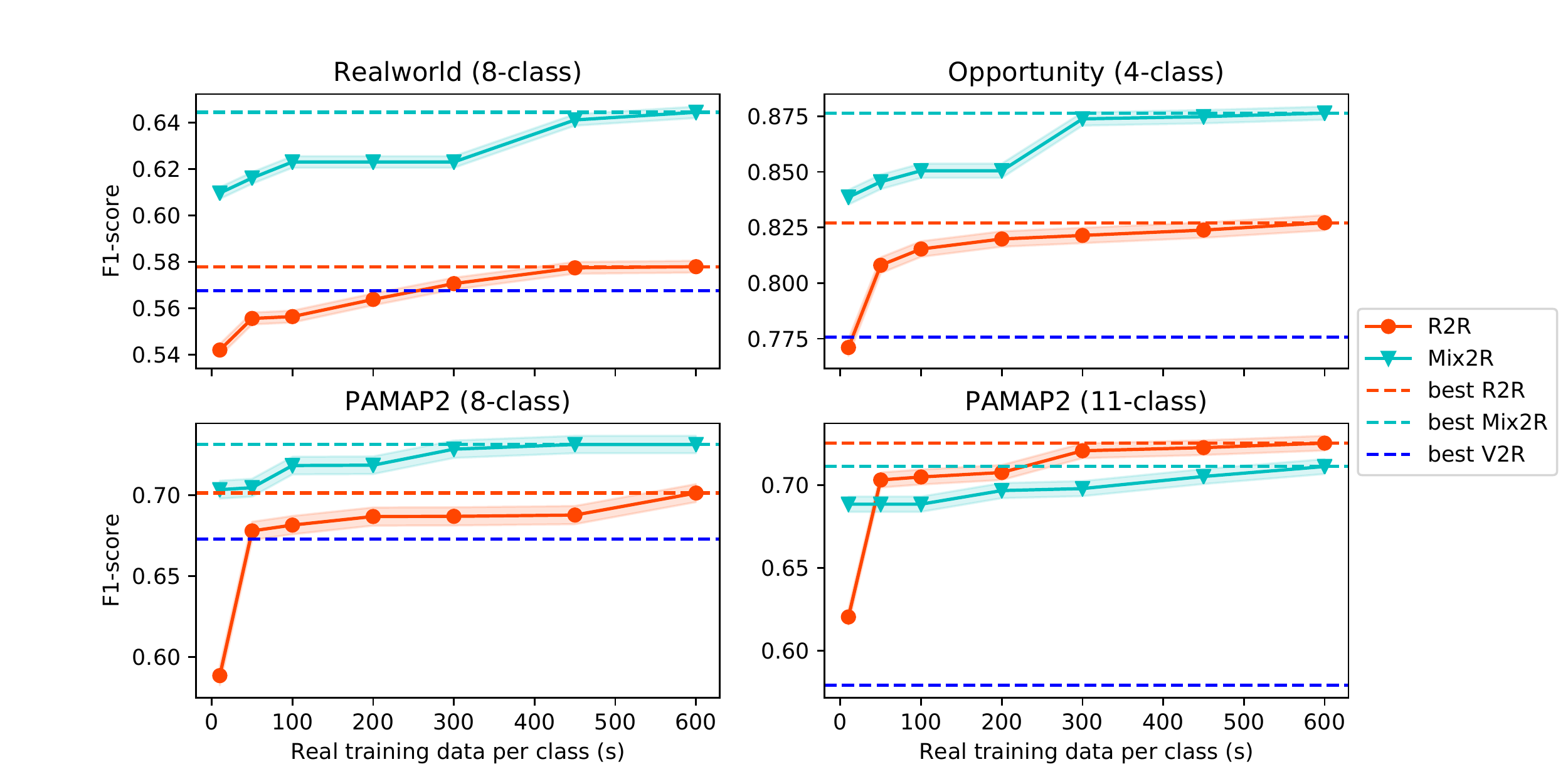}
\caption{\small {Mix2R and R2R performance of a random forest model on 4 different HAR tasks when different amounts of real data per class (in seconds) are used for training. The ratio of virtual data and real data is kept at 1:1 at all datapoints. } }
\label{fig:mix2r}
\vspace*{-1em}
\end{figure}


\subsubsection*{Method} \, {Our first experiment compares the F1-scores achieved by models trained with a mixture of real and virtual IMU data (fixed at 1:1 ratio) as the amount of training data is varied. We repeat this on all 4 datasets and plot the respective learning curves to inspect if the performance gain by Mix2R over R2R is consistent.
Our second experiment compares the F1-score achieved by models trained with a mixture of real and virtual IMU data, where the real IMU data is fixed at 300 seconds per class, but real-to-virtual data ratio is varied from 1:1 to 1:10.}


\subsubsection*{Analysis} \, 
\autoref{fig:mix2r} shows the learning curves on each dataset as the amount of training data is varied. 
{For $3$ out of $4$ datasets, Mix2R outperforms R2R consistently by a clear margin at every inspected point of the learning curves. The greatest performance gain is observed throughout the Realworld learning curve, with an increase of at least 9\% in F1-score by Mix2R over R2R.
Similar trajectories are observed on Opportunity and PAMAP2 (8-class), with the most significant difference between Mix2R and R2R occurring when very limited training data are available (under 100 seconds per class).}

{On PAMAP2 (11-class), Mix2R outperforms R2R when there are only 10 samples per class, and both Mix2R and R2R curves plateau when there are more data available. Given that PAMAP2 (11-class) is also the case where we predict complex activities under the most dissimilar settings, the plot highlights the difficulty of the classification task for both real and virtual IMU data.}

Although it may appear that the learning performance saturates with relatively small amount of data per class (600 seconds per class, for instance) -- we highlight that this has been commonly observed in the literature for the HAR datasets we used (Opportunity, PAMAP2, e.g., \cite{haresamudram2019role, zhao2011cross, reiss2013personalized}). 


{Next, we evaluate the performance achieved when varying virtual and real IMU data mixtures, as presented in \autoref{tbl:multiples}. When compared to the F1-scores of models only trained from real data, adding virtual data to training is seen to give a better or comparable performance at all considered ratios on Realworld, Opportunity, and PAMAP2 (8-class). On Realworld, the greatest gain is seen at 1:5, where the F1-score is increased by 9\% over that at 1:0; At the ratio 1:1, both Opportunity and PAMAP2 (8-class) see improvements of 6$\%$. The performance however does not increase monotonically with the addition of more virtual data. This shows that the effect of mixing virtual and real data is not straightforward. It is possible that as more virtual IMU data is used, the domain shift issue becomes severe and the Random Forest classifier starts to overfit to the virtual IMU data. Adding virtual data has a detrimental effect on PAMAP2 (11-class). This follows our observations in \autoref{fig:mix2r} and can be similarly explained by PAMAP2 (11-class) containing complex activities under the most dissimilar settings in comparison to the virtual IMU data.}


\begin{table}[t]
	\centering
		\caption{\small {Recognition results (mean F1-score, Random Forest classifier) on all datasets.
		Different amounts of virtual data are added to a constant amount of real data, given in seconds per class.}} \label{tbl:multiples}
	\begin{tabular}{c||c||c|c|c|c|c}
		\hline
		Real & Virtual & Real :& & & {PAMAP2}  & {PAMAP2} \\
		Data  & Data & Virtual & {RealWorld} & {Opportunity} & {(8-class)} & {(11-class)} \\
		\hline\hline
		300 & 0    & 1:0  & 0.5706$\pm0.0025$ & 0.8214$\pm0.0034$ & 0.6869$\pm0.0056$ & 0.7206$\pm0.0044$ \\
		300 & 300  & 1:1  & 0.6230$\pm0.0025$ & 0.8738$\pm0.0029$ & 0.7284$\pm0.0053$ & 0.6978$\pm0.0045$ \\
		300 & 600  & 1:2  & 0.6146$\pm0.0025$ & 0.8637$\pm0.0030$ 
		& 0.7006$\pm0.0055$ & 0.7051$\pm0.0045$ \\
		300 & 1500 & 1:5  & 0.6247$\pm0.0024$ & 0.8503$\pm0.0032$
		& 0.6926$\pm0.0055$ & 0.6898$\pm0.0045$ \\
		300 & 3000 & 1:10  & 0.6061$\pm0.0025$ & 0.8396$\pm0.0031$ & 0.6792$\pm0.0056$ & 0.6824$\pm0.0046$ \\
	
		
		\hline\hline
		
	\end{tabular}
	\vspace*{-1em}
\end{table}


Hence, we suggest finding the right balance between the amount of real and virtual IMU data for a model to learn the target activity pattern coexisting in both real and virtual IMU data, before overfitting to the virtual IMU data. We also anticipate that as the quality of virtual IMU improves in future versions of IMUTube, that larger amounts of it will be able to be successfully integrated during HAR training.

\section{TRANSFER LEARNING WITH VIRTUAL IMU DATA FOR HAR CLASSIFIERS} \label{sec:exp3}
In the previous two sections, we have demonstrated that sensor-based human activity classifiers can learn from virtual IMU data, although limitations still exist. So far, we have assumed that labeled virtual and real IMU datasets for target activities are always available. In practice, such a scenario may not always be possible. For example, curating video datasets for virtual IMU data could be challenging, as titles or descriptions of videos can be arbitrarily ambiguous. 

Here, we explore two additional cases for utilizing virtual IMU data: \textit{i)} when the virtual IMU dataset contains a subset of target activity labels; \textit{ii)} when labels for virtual IMU are not available at all. To do so, we leverage two transfer learning setups, \emph{supervised} and \emph{unsupervised}, respectively. 
%
%
{The analysis in this section represents our first attempts in utilizing more sophisticated modeling techniques from deep learning to extend the contribution of IMUTube. Our results are a first step towards handling realistic issues in label collection}, as we do not yet incorporate any automated video labeling or search mechanisms. All experiments follow the same hold-out evaluation protocol detailed in Section~\ref{sec:exp1}.

\subsection{Supervised Transfer Learning}
With supervised transfer learning, we pre-train a model using labeled virtual IMU data and fine-tune it using labeled real IMU data. Importantly, the labels for pre-training and fine-tuning need not match. 
We first explore the setup where virtual and real IMU data share the same set of activity labels -- Imagine we have already curated video and virtual IMU data for some targeted activities, and also collected a small amount of real IMU data; instead of waiting until sufficient amounts of real IMU data is collected, we can first train a model on the virtual IMU data and fine-tune on the small-scale real IMU data. By studying this scenario, we can also gauge if pre-training with virtual data might provide any benefits to activity recognition performance.

Next, we consider a scenario where the virtual IMU data only contains a subset of the real IMU data activity classes. To examine this, we pre-train a model on the virtual PAMAP locomotion (8-classes) task and fine-tune it on the complex activities (11-classes) tasks.




\begin{table}[t!]
	\centering
		\caption{\small Recognition results (mean F1-score) of transfer learning setups when evaluated on different HAR tasks. R2R is the baseline trained on real data from scratch. Transfer learning (TL) results show the performance of the models fine-tuned on real data. } \label{tbl:tl}
	\addtolength{\leftskip} {-0.5cm}
	\begin{tabular}{c|c||c|c||c|c}
		 & & \multicolumn{2}{c||}{DeepConvLSTM} & \multicolumn{2}{c}{CAE+RF} \\
		{Pre-training} & {Fine-tuning} & Supervised R2R & Supervised TL & Unsupervised R2R & Unsupervised TL \\
		\hline\hline
		Realworld & Realworld  & 0.7305$\pm0.0073$ & 0.8337$\pm0.0061$ & 0.7923$\pm0.0067$& 0.7718$\pm0.0069$\\
		Opportunity & Opportunity & 0.8871$\pm0.0074$  & 0.9100$\pm0.0067$ & 0.8896$\pm0.0074$& 0.8477$\pm0.0084$ \\
		PAMAP2 (8-class) & PAMAP2 (8-class) & 0.7002$\pm0.0161$  & 0.7137$\pm0.0159$ & 0.6471$\pm0.0168$& 0.6809$\pm0.0164$ \\
		PAMAP2 (11-class) & PAMAP2 (11-class)  & 0.6977$\pm0.0129$  & 0.7023$\pm0.0129$ & 0.7004$\pm 0.0129$& 0.6989$\pm0.0129$ \\
		PAMAP2 (8-class) & PAMAP2 (11-class)  & -  & 0.7071$\pm0.0129$ & - & - \\
		\hline\hline
	\end{tabular}
\end{table}

\subsubsection*{Method} \, We compare the recognition performance of DeepConvLSTM models \textit{i)} trained only with real data and \textit{ii)} pre-trained on virtual and fine-tuned on real IMU data.
The former is the same as the R2R case in Section~\ref{sec:exp1}. For \textit{ii)}, we randomly split the virtual IMU data into train/validation/test ($80\%/10\%/10\%$) and pre-train the network using the virtual IMU training data. During fine-tuning, all model weights are updated and we report the performance on the real IMU test dataset; In the case where real and virtual IMU activity labels do not match, we replace the last layer of the pre-trained model with the target number of classes (thereby going from 8 to 11 activity classes for PAMAP2) and update all the network weights. We present results on all 4 datasets and follow the training and evaluation protocols described in Section~\ref{sec:exp1}.

\subsubsection*{Results} 
\begin{figure}[t]
\centering
\vspace*{-1em}
\includegraphics[width=\textwidth]{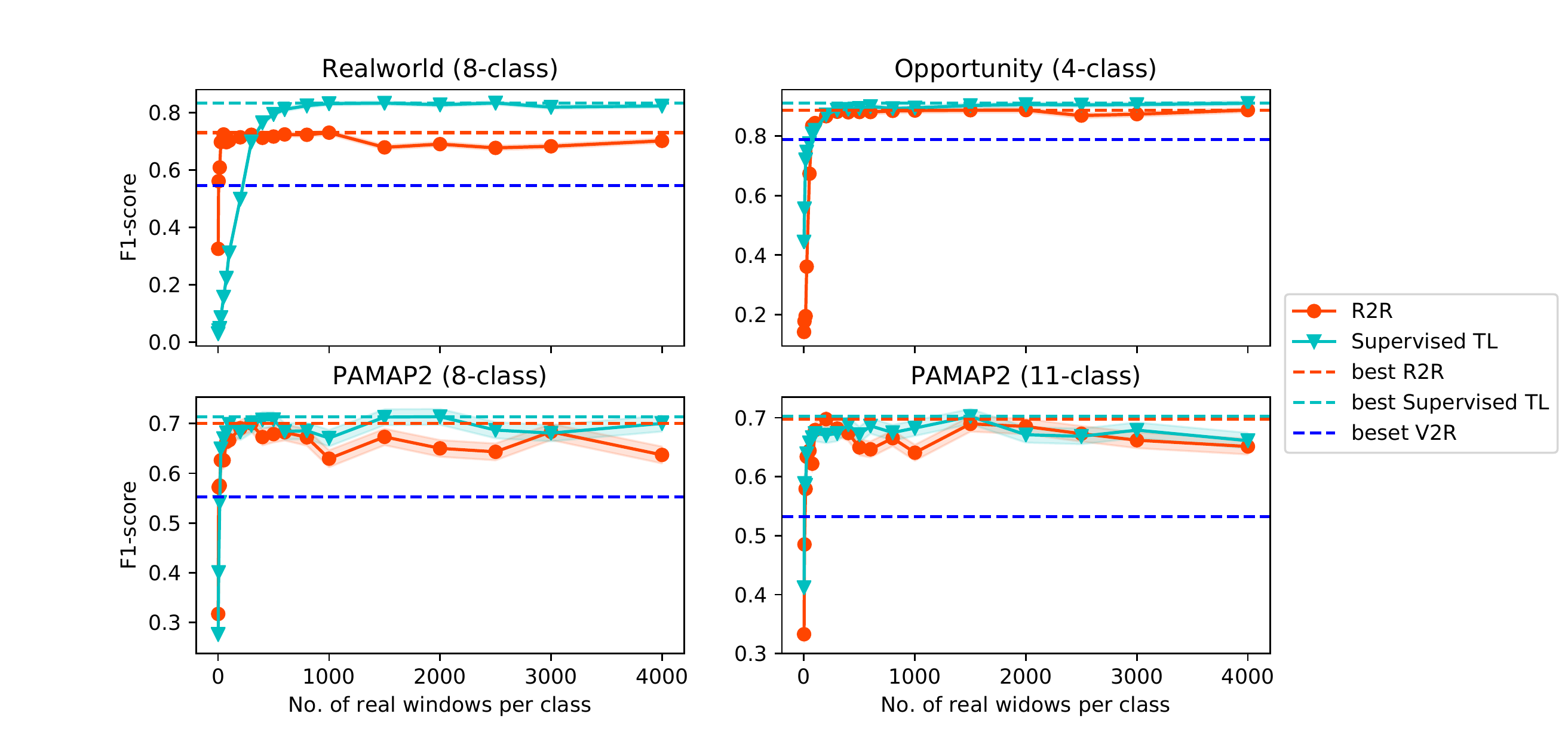}
\caption{\small Transfer learning vs. amount of real data used for training.}  
\label{fig:dl_tl}
\vspace*{-1em}
\end{figure}

\, The left of \autoref{tbl:tl} shows the differences in F1-scores achieved by DeepConvLSTM models with (Supervised TL) and without pre-training (Supervised R2R) when evaluated on a random test subject. With pre-training, statistically significant performance gains are observed on Realworld and Opportunity, at 14\% and 3\% respectively. We further present the effects of using different amounts of real IMU data for fine-tuning the selected base model in \autoref{fig:dl_tl}. We see the most obvious difference in learning trajectories on the Realworld dataset, where only a small amount of real data is needed to fine-tune the base model such that it surpasses R2R performance. 
{In the last row of \autoref{tbl:tl}, we also show results for transfer learning in the case where virtual and real IMU data labels do not match (PAMAP2 8-class for pre-training, PAMAP2 11-class for fine-tuning). In this case, the model with pre-training achieves an F1-score of 0.71 on PAMAP2 (11-class), which is statistically comparable to the result in the R2R setting.}

\subsection{Unsupervised Transfer Learning}
Unsupervised transfer learning considers the scenario where we extract the virtual IMU data from a large body of videos without labels. Curating a collection of unlabeled videos is easier relative to obtaining labeled videos, particularly in scenarios where the video descriptions/labels may be unreliable. Without a set of specific target activities in mind, any videos with humans can be utilized. 
{Validating the feasibility of this approach represents a first step towards curating a large collection of virtual IMU data, consisting of very diverse movements and activities from which a model can learn generic representations.}



\subsubsection*{Method} \,
Our unsupervised transfer learning setup consists of two stages. The first stage pre-trains a convolutional autoencoder (CAE) to learn feature representations. The second stage extracts from real IMU data the learned representations, which are then used to train a random forest classifier. 
{We compare the recognition performance achieved by the entire CAE-RF setup when: \textit{i)} virtual IMU data is used for training the CAE (Unsupervised TL); and \textit{ii)} when real IMU data is used for training the CAE (Unsupervised R2R).}

We use \citeauthor{haresamudram2019role}'s architecture, where the encoder contains four convolutional blocks, leading to the bottleneck layer~\cite{haresamudram2019role}. Each block contains two $3$x$3$ convolutional layers followed by $2$x$2$ max-pooling. Batch normalization is applied after each layer \cite{ioffe2015batch}. The output from the last convolutional block is flattened before being connected to the bottleneck layer. The decoder inverts the encoder by performing convolution, interpolation, and padding to match the sizes of the corresponding encoder blocks \cite{odena2016deconvolution}. ReLU activation \cite{nair2010rectified} is used throughout, except the output, where the hyperbolic tangent function is used instead. We follow the evaluation protocol used in the DeepConvLSTM case.



\subsubsection*{Results} \, 
The right part of Table~\ref{tbl:tl}, shows results for using virtual IMU for pre-training with varying performance relative to models trained on real IMU data. {On Realworld, Opportunity and PAMAP2 (11-class), unsupervised TL using virtual IMU data reached up to $95\%-100\%$ of R2R F1-scores. On PAMAP2 (8-class), we even see an increase of 5\% over the R2R protocol. These results demonstrate the feasibility in utilizing virtual IMU data even in scenarios where video labels are completely absent.}

\section{DISCUSSION} \label{sec:discussion}
In this section, we discuss the implications of the results presented, limitations in our approach, and highlight opportunities which this work opens up.

\subsection{{Demonstrating Feasibility}}

We have presented a processing pipeline and a series of validation studies to support our thesis that an automated pipeline from video to virtual IMU data can replace the labor-intensive practice of collecting labeled datasets from real on-body IMU devices. 
{
IMUtube shows how a full three-axis virtual accelerometer sensor derived from arbitrary videos can be utilized for human activity recognition.} 
The automated pipeline provides the opportunity to collect much larger labeled data sets, which in turn can improve classifiers for human activity recognition.




{Our validation experiments explored ways to model virtual IMU data, either standalone or in conjunction with real IMU data. On three different datasets (Realworld, Opportunity and PAMAP2 8-class), training from virtual IMU alone led to competitive results compared to those from real IMU data (recovering up to 90\%), and a simple mixing of data from two sources brought considerable gains (4\%-12\% increase) to recognition performance.}

PAMAP2 11-class is a special case as the activity recognition task extends to complex, non-locomotion human activities, such as vacuum cleaning. It is also different because we have utilized a diverse range of video data collected over multiple visual datasets. Our results show that we can indeed still learn from virtual data under such settings, and our V2R results still reach at least {80\%} when compared to R2R (\autoref{tbl:p11_rf}). However, what is still missing is that we have not seen an improvement over R2R results through mixing (2\% decrease) or transfer learning (insignificant change). While this suggests that modeling complex activities and mixed data sources remain issues, we believe that modelling complex activities, and by extension, the merit of using more accelerometry data (be it virtual or real) still warrants further investigation.  For example, is there simply an upper bound to predicting these complex activities using motion-based data alone? 






\subsection{Limitations and Extensions of Current Approach} \label{sec:limitations}
We have utilized a series of off-the-shelf techniques at every step of the proposed pipeline in Section \ref{sec:processing-pipeline}. While this supports reproducibility of our results, it does result in limitations that impact the overall quality of labeled data for HAR. We discuss the known limitations of each step of the pipeline and present steps forward to advance this line of research. 

\subsubsection{From Vision To Pose} \, Accurate recovery of the human skeleton pose from videos has known limitations arising from the movement of both the subjects as well as the camera.

{
The 2D pose estimator used in IMUTube, \emph{OpenPose}, has previously known failure scenarios including partial detection of joints, swapping between left and right for rare poses, self-occlusion from the camera viewpoints, and partially visible bodies~\cite{cao2017realtime}.
Such errors in the estimated 2D pose propagate to the 3D pose estimation, which itself is a challenging problem due to the inherent uncertainty of the added third dimension~\cite{pavllo20193d}.
Erroneous 2D and 3D poses may distort corresponding perspective projections (PnP) between the poses leading to wrong 3D pose calibration~\cite{Hesch_2011}.
Depth map or camera ego-motion estimation for a dynamic scene can be imprecise when having occlusions or motion blur between foreground and background objects, or when light condition changes~\cite{Gordon_2019}.
Therefore, the current pipeline can result in distorted 3D motion due to the accumulated errors, since these challenges are common for videos in the wild.
}

For the recovery of the human skeleton pose, we would expect improvements based on solutions that leverage more sophisticated pose tracking techniques that are more robust to vigorous movement, a change of scenery, the presence of multiple people, and occlusion. 
Also, camera movements relative to the people captured in the videos could come from the instability of the camera (e.g., for hand-held cameras) as well as video filming techniques (e.g., panning shots).  Specialized video stabilization strategies or camera ego-motion techniques can address these issues~\cite{shen2019human,yu2019robust,zhou2017unsupervised}.
We believe that the application of these techniques (and others not yet mentioned or even developed) will further improve pose extraction quality and expand the variety of videos that can be treated as input to our pipeline.
\subsubsection{From Pose To Accelerometry} \, Our current approach assumes an equivalence between acceleration measured by a device on the wearer's body with that measured at the nearest body joint. This view discounts any consideration of factors such as body mass, device movement and skin friction. To better model the on-body location of IMU devices, utilizing techniques from body mesh modeling is a straightforward solution to increase realism to the pipeline. We foresee that investigating the use of body mesh might also bring up the possibilities of synthesizing credible accelerometry data from people of different body shapes from the movement of a single human skeleton pose~\cite{pons2015dyna,kanazawa2018end,loper2015smpl}. In addition, while we have only considered the generation of virtual accelerometry data in this work, we can adapt most parts of the pipeline to generate the full set of IMU signals, including gyroscope and magnetometer readings.


\subsubsection{From Accelerometry to Virtual IMU} \, Real IMU data, which have been the basis of building HAR classifiers, are not free of noise. Sensor noise may come from factors such as drift, hysteresis and device calibration. To carry over such characteristic sensor noise on our virtual data, domain adaptation techniques can be deployed as well as more  sophisticated techniques like Generative Adversarial Networks.~\cite{rosca2018distribution,goodfellow2014generative} 

\subsubsection{Learning from Virtual Data} \, A domain shift exists when a machine learning model is trained from virtual data and tested on real IMU data. Again, domain adaptation strategies to the input of the machine learning model is a solution. Alternatively, it will be promising to investigate domain-invariant features learned from virtual and real data, which could potentially lead to performance gains in HAR. 

\subsubsection{Which Videos are Currently Suitable for IMUTube?} \, 
{Our qualitative inspection of the IMUTube output and the per-class V2R results suggests that certain video features may contribute to poorer recognition performance on virtual IMU data: large ego-motions, multiple moving objects and people, and occlusion.}

{Large ego-motions can be found in the Realworld `running' videos in which the video-taker was also running, leading to significant vertical shaking motion from the camera itself. It is possible that such vertical motions end up producing features that are very similar to those from a `jumping' motion, which may explain a higher class confusion observed between `running' and `jumping' on the Realworld and PAMAP2 (8-class) tasks. An additional factor might have come from the presence of multiple moving objects and people in the background (e.g. pedestrians) in the `running' videos.}

{We also found that V2R models struggle more in classifying activities with similar poses which have more subtle differences in limb movements, e.g. `standing` vs. `vacuum cleaning`, `sitting' vs. `ironing', as in PAMAP2 (11-class). In many `vacuum cleaning' and `ironing' videos, the subject's arm movements are occluded by objects in the scene, e.g. clothes or home furniture.}

{On the other hand, videos with fewer or without such motion artifacts tend to produce virtual IMU data that are well-classified under V2R. Moreover, videos featuring activities with distinctive poses and motions, e.g. cycling, are well-classified under the V2R setting. There are also many existing techniques that will allow us to further tackle motion blur (\cite{shi2014discriminative,alireza2017spatially}) and occlusion (\cite{girshick2015fast,redmon2016you}). It is also possible that the future curation of video data can automatically rank videos by the presence of these undesirable features to arrive at a suitable dataset for virtual IMU data extraction.} 

\subsection{The Road Ahead}

Our primary goal in this paper was to motivate the HAR community with a promising approach that overcomes the main impediment to progress---lacking large labeled data sets of IMU data. While technical challenges remain, we have validated this approach and provide a processing pipeline that the community can collectively develop. Here we highlight the most compelling research opportunities.


\subsubsection{Large-scale Data Collection} \, The ultimate goal, as suggested by the name of IMUTube for our initial tool, is to develop a fully automated pipeline that begins with the retrieval of videos representing particular human activities from readily available sources (e.g., YouTube) and converts that video data to labeled IMU data. Since it is much more common to have video evidence of the wide variety of human behaviors, this is an obvious advantage over past labor-intensive and small-scale efforts to produce such HAR datasets. We have shown great promise with this direction, and above listed some known limitations that can be addressed by different vision, signal processing, and machine learning techniques. The reader will note that the videos used for our validation studies were also curated, meaning there was a significant effort in selecting appropriate video examples.  The hope is that this curation effort can also be reduced and ultimately eliminated because the sheer number of relevant videos will overcome the deficiencies of less useful video data.


\subsubsection{Deep Learning} \, {Deep Neural Networks have} transformed recognition rates in other fields~\cite{oord2016wavenet,he2016deep}, but HAR has lagged behind, again due to the lack of large corpora of labeled data. While we expect that IMUTube is a significant advance towards that goal, having the data alone is not the end goal. We have not yet produced a large-scale HAR dataset, and until we do so we can only hope that deep learning techniques will take over. We then fully expect HAR to inform deep learning techniques. 

\subsubsection{Extending the Field of HAR} \, An important advantage to generating virtual IMU data is that we can place the virtual sensor in a wide variety of places on the human body.  While some of the standard datasets we used in this work have subjects wearing multiple IMUs, there are limits to how many devices one can wear and still perform activities naturally.  IMUTube removes that limitation. Now, for any given activity, we can experimentally determine where to place one IMU (or multiple IMUs) to best recognize that activity.  For a set of activities, which place optimizes the recognition of all of the activities in that set.  We have never had the ability to contemplate that kind of question. 
We also need not limit to IMUs placed directly on the body.  Models of how clothing responds on a body might be used to generate virtual IMU data for objects that are loosely connected to the body~\cite{pons2017clothcap,alldieck2019learning}.
HAR can now inform clothing manufacturers of where in the material for a shirt, for example, one would want to integrate IMU data collection to predict the activities of the person wearing the shirt, or any other piece of clothing for that matter~\cite{kang2017development,muhammad2020review}.


\subsubsection{Real IMU as `Seeds' to Our Pipeline} \, While IMUTube is about generating lots of virtual IMU data, our results show the value for the more traditional curated datasets from real IMU data.  The real IMU data provides a seed that the virtual data grows into more sophisticated HAR models.  
Now the efforts in real IMU data collection can be focused on producing very high quality labeled data from a wide enough variety of subjects performing key activities. It may even be the case that this real IMU seed data is the treasured commodity that companies can use to provide the best seeds for IMUTube-generated virtual IMU data and the models grown from them. 
\section{RELATED WORK} \label{sec:related work}

The proposed method details a pipeline towards opportunistically extracting virtual sensor data from a potentially very large body of publicly available videos. 
This is in contrast to current wearable sensor data collection protocols, which involve user studies and human participants, as well as other approaches that generate sensor data from motion capture (mocap) settings. 
In what follows, we first discuss approaches to data collection for sensor-based human activity recognition as well as mocap based techniques. 
These approaches represent the state-of-the-art in the field that are based on dedicated data recording protocols.
Subsequently, we detail prior work on training classifiers with limited labeled data, thereby focusing on data augmentation techniques and transfer learning. 

\subsection{Sensor Data Collection in HAR}
Sensor data collection for human activity recognition is often performed by conducting user studies~\cite{Chavarriaga13prl,Zhang12ubicomp,reiss2012introducing}. 
Typically, the participants in a study are asked to perform activities in laboratory settings while wearing a sensing platform.
The advantage of data recording in a lab setting is that in addition to sensor data typically video data is recorded that is subsequently used for manual data annotation.
For this purpose, the sensor and video data streams need to be synchronized \cite{plotz2012automatic}, and human annotators need to be trained for consistency in annotation. 
The laboratory is designed to resemble a real-world environment, and user activities are either scripted or naturalistic. 
These include various gesture and locomotion level activities. 
However, designing a lab study to capture realistic natural behaviors is difficult. 
The protocol of such studies makes it challenging to collect large scale datasets. Furthermore, the annotation of activities is costly and error-prone and therefore prohibitive towards creating large datasets as they are required for deriving complex machine learning models. 

Recently, Ecological Momentary Assessment (EMA) based approaches have been employed to record and especially annotate real-world activity data \cite{laput2019sensing,thomaz2015practical,hovsepian2015cstress}. 
The sensing apparatus (containing sensors such as accelerometers or full-fledged IMUs) is worn on-body, and users self-report the activity labels when they are asked to do so through direct notification.
Although these methods may lose sample-precise annotation of the activities, they encourage the collection of larger-scale datasets. 
While limited to gesture-based activities, Laput and Harrison \cite{laput2019sensing} have shown that larger numbers ($83$) of fine-grained hand activities can be reliably recorded and annotated.
Both in-lab and EMA based collection protocols directly involve human participants to collect movement data using body-worn sensors. 

Other approaches have explored alternative data collection methods that do not directly involve human participants.
Kang \etal\ render a $3$D human model on computer graphics software 
and simulate human activities \cite{kang2019towards}. 
The sensor data is extracted from the simulated human motion, and subsequently used to train the recognition models. 
However, it is very difficult to realistically simulate and design complex human activities. 
Therefore, such methods typically only explore simple gestures and locomotion activities. 
{Alternatively, \cite{xiao2020deep,takeda2018multi} extract sensory data from public, large-scale motion capture (mocap) datasets~\cite{mahmood2019amass,cmumocap,ofli2013berkeley}, which contain a variety of motions and poses for human activity recognition. 
Although these datasets cover hundreds of subjects and thousands of poses and motions, they rarely include everyday activities. 
The majority of such mocap datasets include dancing, quick locomotion transitions, and martial arts, which are less relevant to recognizing daily human activities.}

{Most related to our work, Rey~\etal~\cite{rey2019let} also proposed to collect virtual sensor data from online videos and demonstrated the effectiveness of the virtual sensor data for recognizing fitness activities.
Their approach computes the 2D pose motion for a single person in the video with a fixed camera viewpoint.
A regressor is trained for a target real sensor with the synced video and accelerometer recordings, which transfers the changes in joint locations from the 2D scene to the norm of the three-axis accelerometer.
In contrast, our work can generate data from the full IMU (three-axis accelerometer, gyroscope, and simulated magnetometer). Further, we perform 3D motion estimation from videos with multiple people and scenes in the wild using camera motion tracking. 
We do not require synced video and wearable recordings as the virtual sensor can be adapted to any real sensor with our efficient distribution mapping method.}

We leverage the availability of large scale video datasets that cover real-world activities to extract sensory data. 
These videos are recorded in-the-wild and contain a wide range of activities, including everyday activities, which makes them very attractive for deriving realistic and robust human activity recognition systems.


\subsection{Tackling the Sparse Data Problem}
Many publicly available datasets for human activity recognition contain imbalanced classes. For example, approximately $75\%$ of the Opportunity dataset (which has $18$ classes in total) \cite{Chavarriaga13prl} consists of the null class \cite{guan2017ensembles}, making it challenging to design classifiers. The activities being studied also impact the class imbalance to some extent. In the PAMAP2 dataset, the skipping rope class constitutes approximately $2.5\%$ of data, relative to other activities which constitute around $9\%$ on average \cite{guan2017ensembles}. 
This follows reason as, unless your name is Rocky Balboa, 
it is harder for subjects to perform rope skipping for longer durations of time, in contrast to walking or lying down. 
This resulting class imbalance poses a challenge for the design and training of classifiers, which may find it easier to simply predict the majority class. 
Furthermore, the relatively small size of labeled datasets results in models quickly overfitting and does not allow the application of complex model architectures. It is also difficult to apply potentially alleviating techniques such as transfer learning, which rely on large datasets for knowledge transfer. {As a result, \cite{tong2020are} have noticed that the adoption of deep learning methods in human activity recognition has not yet translated to the pronounced accuracy gains seen in other domains.}

As a way to overcome the problem of small, class-imbalanced datasets, data augmentation techniques have been applied previously to prevent overfitting, improve generalizability and increase variability in the datasets. 
They involve techniques that systematically transform the data during the training process in order to make classifiers more robust to noise and other variations \cite{mathur2018using}. 
They artificially inflate the training data by utilizing methods, which perform data warping, or oversampling \cite{shorten2019survey}. 
Data warping includes geometric transformations such as rotations, and cropping, as well as adversarial training. 
For time series classification, the data warping techniques include window slicing, window warping, rotations, permutations and dynamic time warping \cite{le2016data,fawaz2018data}. 
Several of these transformations can be combined to further improve the performance over a single method. 
Um~\etal~demonstrate that combining three basic methods (permutation, rotation and time warping) yields better performance than using a single method \cite{um2017data}. 
In \cite{rashid2019times}, construction equipment activity recognition is also improved by combining simple transformations. 

{Recently, data generation using either oversampling or generative adversarial networks (GANs \cite{shorten2019survey}) have also been successfully introduced to sensor-based human activity recognition \cite{yao2018sensegan}. 
However, in contrast to other domains such as computer vision, performance improvements remain moderate, most likely due to non-trivial challenges inherent to generating realistic yet novel timeseries data.
Oversampling based methods include synthetic minority oversampling technique (SMOTE) \cite{ fernandez2018smote}.
GANs have been used to, for example, augment biosignals \cite{ haradal2018biosignal} or in IoT \cite{yao2018sensegan}.
Extending the conventional GAN approach, in \cite{ramponi2018t}, a data augmentation technique for time series data with irregular sampling is proposed utilizing conditional GANs. 
It is shown to outperform data warping techniques such as window slicing and time warping. }
Augmentation for wearable sensor data has been explored for monitoring Parkinson's disease in \cite{um2017data}. 
In this paper, seven transformations, including jittering, scaling, rotation and warping are detailed and their effects relative to no augmentation is studied. 
Further, the authors observed that combining multiple transformations results in higher performance. 
In \cite{steven2018feature}, augmentation is performed on IMU spectrogram features to improve the activity recognition performance. 

Another approach to deal with small labeled datasets includes transfer learning. 
Here, a base classifier (typically a neural network) is first trained on a base dataset and task. 
Subsequently, the learned features are re-purposed, or \emph{transferred}, to a second target network to be trained on the target dataset and task. 
In particular, if the target dataset is significantly smaller compared to the base dataset, transfer learning enables  training a large target network without overfitting \cite{yosinski2014transferable}, and typically results in improved performance. 
In \cite{saeed2019multi}, the authors propose a self-supervision pretext task and demonstrate its effectiveness for unsupervised transfer learning on other datasets with little labeled data. 
A more extreme example of having very small labeled datasets includes one-shot and few-shot learning, which contain very few labeled samples per class \cite{feng2019few}.

While the data augmentation techniques do improve the classification performance, they, ultimately, produce perturbed training samples. 
Therefore, they are unable to provide for the variety in human movements that is obtained by collecting data from a large number of subjects. 
{On the other hand, the GAN based techniques perform augmentation by sampling from the dataset distribution. 
However, they require substantial amounts of data to train, and may suffer from training instability and non-convergence \cite{yao2018sensegan}. 
Furthermore, there is limited prior work studying data augmentation by GANs for wearable sensor data and their actual suitability for sensor-based human activity recognition remains to be shown.
This makes it challenging to readily apply these generative networks to create more data. }

We tackle the problem of having small labeled datasets with a different approach -- by generating large quantities of virtual IMU data from videos. 
As we can leverage a large body of videos, containing many individuals, we generate datasets containing more diverse movements and potentially much larger datasets of realistic data, which is in stark contrast to existing methods that try to combat the sparse data problem.

\section{CONCLUSION} 

In this paper we {developed a framework for}  generating virtual IMU data based on automated extraction from video as a means to collect large-scale labeled datasets to support research in human activity recognition (HAR).  
We designed and validated {our framework}, IMUTube, that integrates a collection of techniques from computer vision, signal processing, and machine learning. 
Our initial findings show great promise for this technique to extend the capabilities for HAR, {at a minimum for simple activities whose main IMU characteristics are confined to expression in 2D}. 

The greater promise of this work requires a collective approach by computer vision, signal processing, and activity recognition communities (who have already been greatly united through the advances of deep learning) to advance {the underlying agenda}.  
Computer vision researchers can clearly build upon the IMUTube pipeline to address a variety of current limitations, further automating the pipeline and reducing the need for human curation of online videos. 
Signal processing advances can further manipulate the virtually-generated data to better condition the virtual data and represent the features and distributions of real IMU data. 
Activity recognition researchers can apply known modern learning techniques to this new class of labeled data for HAR and develop more effective ways to model, both with and without a mixture of real IMU data. 
Within a few years, we expect this collective effort to result in HAR as yet another success story for large-data-inspired learning techniques.



\bibliographystyle{ACM-Reference-Format}
\bibliography{./bibs/strings,./bibs/har,./bibs/dl,./bibs/virtual,./bibs/mocap,./bibs/pose2D,./bibs/pose3D,./bibs/pnp,./bibs/extrinsic,./bibs/intrinsic,./bibs/icp,./bibs/depth,./bibs/sbd,./bibs/sparse_data,./bibs/tracking,./bibs/stats,./bibs/vision,./bibs/nlp,./bibs/speech,./bibs/sensor,./bibs/graphics}

\end{document}